\documentclass{article}
\usepackage{iclr2027_conference,times}

\usepackage{amsmath,amsfonts,bm}

\def\eqref#1{equation~\ref{#1}}

\def\1{\bm{1}}

\def\mD{{\bm{D}}}

\def\mH{{\bm{H}}}

\def\mL{{\bm{L}}}

\def\mR{{\bm{R}}}

\def\mW{{\bm{W}}}

\DeclareMathAlphabet{\mathsfit}{\encodingdefault}{\sfdefault}{m}{sl}
\SetMathAlphabet{\mathsfit}{bold}{\encodingdefault}{\sfdefault}{bx}{n}

\renewcommand{\eqref}[1]{\textup{(\ref{#1})}}

\usepackage{algorithm}
\usepackage{algpseudocode}
\usepackage{amsmath,amssymb,amsthm,bm}
\usepackage{booktabs}
\usepackage{graphicx}
\usepackage{subcaption}
\usepackage{microtype}
\usepackage{multirow}
\usepackage[table]{xcolor}
\usepackage{hyperref}
\usepackage{url}
\usepackage{pifont}
\usepackage{etoolbox}
\usepackage{placeins}
\definecolor{revisionred}{RGB}{185,0,0}
\newif\ifreviewchanges
\reviewchangesfalse
\newcommand{\rev}[1]{{\ifreviewchanges\color{revisionred}\fi #1}}
\newcommand{\method}{\textit{H-SVDQuant}}

\title{Deflating the Hessian: Rank-4 W4A4 Quantization for Multimodal Diffusion Transformers}

\hypersetup{
    hidelinks,
    pdftitle={Deflating the Hessian: Rank-4 W4A4 Quantization for Multimodal Diffusion Transformers},
    pdfauthor={Shiwen Wang, Pengxiang Zhao, Xiaoming Yuan},
    pdfkeywords={diffusion transformers, post-training quantization, W4A4 quantization,
    Hessian deflation, low-rank correction, activation quantization}
}

\author{
Shiwen Wang$^{1}$ \quad Pengxiang Zhao$^{2}$ \quad Xiaoming Yuan$^{1,*}$\\[0.4ex]
$^{1}$The University of Hong Kong \qquad $^{2}$Huawei Technologies Co., Ltd.\\[0.4ex]
$^{*}$Correspondence to: \texttt{xmyuan@hku.hk}
}

\iclrfinalcopy
\begin{document}
\maketitle
\raggedbottom

\begin{abstract}
In diffusion transformers, low-rank branches can mitigate 4-bit weight--activation (W4A4) post-training quantization (PTQ) loss by decomposing each weight into a low-bit residual and a high-precision low-rank component.
Existing low-rank PTQ approaches, however, either optimize low-rank compensation and residual quantization separately, often requiring higher ranks, or rely on second-order weight updates without explicitly modeling activation quantization error, which becomes particularly pronounced under 4-bit quantization. 
To address these limitations, we present \method{}, a unified framework modeling low-rank-assisted W4A4 PTQ as a coupled calibration problem and deriving optimization-based solvers from the joint objective. Eliminating the output-side low-rank factor yields a \emph{deflated Hessian} that discounts residual errors already captured by the low-rank component, while an activation-noise surrogate is incorporated to suppress activation quantization error. 
Across five diffusion backbones, rank-4 \method{} consistently outperforms rank-4 SVDQuant in PSNR and LPIPS. It further surpasses rank-32 SVDQuant on SANA-1.6B, FLUX.1-schnell, and FLUX.1-dev with an $8\times$ smaller rank and up to $6.25\times$ faster quantization. Furthermore, on the Qwen3-8B LLM, rank-4 \method{} improves MMLU accuracy from 61.50\% to 68.17\% over rank-32 SVDQuant. Overall, \method{} achieves better W4A4 performance with substantially lower rank and quantization cost.
\end{abstract}
\section{Introduction}

Diffusion transformers (DiTs)~\citep{peebles2023dit} repeatedly apply large linear transformations along the denoising trajectory, making low-precision inference an attractive route to reducing memory traffic and arithmetic cost. Weight--activation 4-bit post-training quantization (W4A4 PTQ) reduces both weights and activations to low precision without retraining, but the resulting errors can accumulate across layers and amplify over denoising steps. 


SVDQuant~\citep{li2025svdquant} first extracts a truncated-SVD branch and then quantizes the remaining residual. LRC~\citep{scetbon2024lrc} jointly optimizes a quantized weights and low-rank correction. LoRaQ~\citep{bouquet2026loraq} directly optimizes a low-rank perturbation to compensate for weight quantization error. GPTQ-intrinsic LoRA~\citep{zhang2026gptqintrinsic} also incorporates low-rank correction into second-order quantization.
Despite these advances, existing methods do not explicitly formulate a multivariable coupled optimization problem, nor derive an efficient algorithm that jointly captures weight- and activation-side quantization under a small rank budget.
This leads to our research question: \emph{
how can we jointly model the variables governing low-rank-assisted W4A4 PTQ and solve the resulting problem efficiently to maximize the effectiveness of low-rank branches?
}

\begin{figure}[t]
\centering
\begin{subfigure}[b]{0.35\linewidth}
    \centering
    \includegraphics[width=\linewidth]{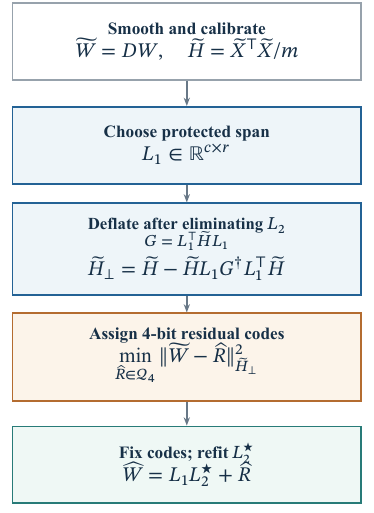}
    \caption{Hessian deflation metric.}
    \label{fig:method_panel}
\end{subfigure}\hfill
\begin{subfigure}[b]{0.63\linewidth}
    \centering
    \includegraphics[width=\linewidth]{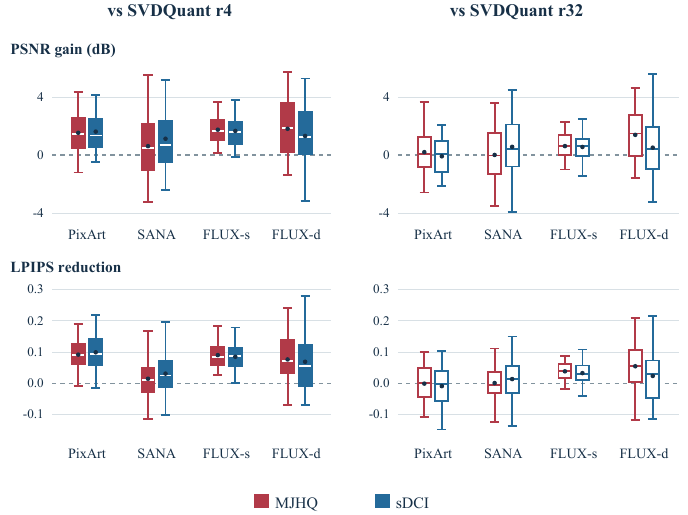}
    \caption{Paired rank efficiency gains over SVDQuant.}
    \label{fig:intro_boxplot}
\end{subfigure}

\caption{\rev{
\textbf{(a) Residual quantization with Hessian deflation:} For a fixed protected subspace $\mL_1$, Hessian deflation ($\widetilde \mH_\perp \mL_1=0$) focuses residual quantization strictly on errors outside the low-rank subspace.
\textbf{(b) Paired rank efficiency:} The paired distributions show consistent gains over rank-4 SVDQuant, and show model-dependent comparisons with rank-32 SVDQuant, with the strongest rank-efficiency gains on SANA and the FLUX variants.}}
\label{fig:intro_method_pair}
\end{figure}

We present \emph{\method{}}, a unified framework for low-rank W4A4 PTQ that formulates a joint reconstruction objective, derives optimization-based solvers, and instantiates them in a practical calibration algorithm.
In Figure~\ref{fig:intro_method_pair}(a), a smoothed weight matrix \rev{$\widetilde \mW=\mL_1\mL_2+\mR$}, where \rev{$\mL_1,\mL_2$} denote the low-rank component and \rev{$\widehat \mR=Q(\mR)$} is the quantized residual.
\rev{$\mL_1$} specifies the protected input subspace with calibration-weighted initialization. Analytically eliminating \rev{$\mL_2$} yields 
\begin{equation}
\mH_\perp
=
\mH-\mH\mL_1\left(\mL_1^\top \mH \mL_1\right)^\dagger \mL_1^\top \mH,
\qquad
\mH_\perp \mL_1=0,
\end{equation}
where \rev{$\mH$} is the calibration activation Gram matrix and $(\cdot)^\dagger$ denotes the Moore--Penrose pseudoinverse. We refer to this reduced geometry as \emph{Hessian deflation}: 
Since \rev{$\mH_\perp \mL_1=0$}, errors that lie in the protected input subspace can be compensated by the low-rank component.
\rev{$\widehat \mR$} therefore only needs to account for the remaining errors measured by $\mH_\perp$. Then \rev{$\mL_2$} is refitted using a closed-form solution $\mL_2^\star$. 
We optimize activation quantization through the diagonal smoothing and the activation-quantization surrogate. Specifically, 
we incorporate a smoothing matrix $\mD$ and derive its \(\ell_p\)-relaxed initialization with geometric programming, leaving the smoothed weights $\widetilde \mW = \mD\mW$ for subsequent quantization, \rev{avoiding costly grid searches or online rotation overhead while accelerating calibration.} Furthermore, our algorithm accommodates an activation-noise surrogate $\lambda_A$ to suppress the amplification of activation quantization error during residual quantization, specified in Section~\ref{sec:method}.

We evaluate \method{} against PTQ baselines for DiTs:  SVDQuant~\citep{li2025svdquant}, DiRotQ~\citep{sharify2026dirotq}, OrbitQuant~\citep{lee2026orbitquant}, and ViDiT-Q~\citep{zhao2025viditq}. Our backbones include diffusion transformers ranging from 0.6B to 20B parameters: PixArt-$\Sigma$~\citep{chen2024pixartsigma}, SANA-1.6B~\citep{xie2024sana}, FLUX.1-schnell~\citep{blackforestlabs2024flux}, FLUX.1-dev~\citep{blackforestlabs2024flux}, and Qwen-Image~\citep{qwenimage2024} (including its variant with 4-step Lightning LoRA). Under W4A4, \method{} consistently improves reconstruction fidelity across all backbones (up to $+1.61$~dB PSNR on PixArt-$\Sigma$) and rivals or surpasses the $8\times$-larger rank-32 SVDQuant baseline on SANA-1.6B and FLUX models (Figure~\ref{fig:intro_method_pair}(b)). Evaluation on Qwen3-8B~\citep{yang2025qwen3technicalreport} also confirms that rank-4 \method{} achieves 10.44 WikiText-2 perplexity (PPL) and 68.17\% MMLU accuracy (outperforming rank-32 SVDQuant at 11.21 PPL and 61.50\% MMLU), suggesting that the proposed formulation also transfers to large language transformers (LLMs).

Our contributions are threefold:
\begin{itemize}\setlength{\itemsep}{1pt}\setlength{\parskip}{0pt}
    \item \textbf{Unified low-rank-assisted W4A4 PTQ problem.} 
    We formulate a joint calibration objective for low-rank-assisted W4A4 PTQ that accounts for both weight- and activation-side quantization, seamlessly integrating an activation quantization surrogate objective. 
    \item \textbf{Efficient optimization-based solvers.}  We derive the optimization-based solvers for our framework, integrating calibration-weighted low-rank initialization, diagonal smoothing, second-order residual quantization, and closed-form low-rank refitting.
    \item \textbf{Extensive empirical validation under small-rank budgets.} We evaluate rank-4 \method{} across five diffusion backbones and Qwen3-8B, surpassing rank-32 SVDQuant on SANA and the FLUX variants while accelerating offline PTQ calibration by up to $6.25\times$.
\end{itemize}

\section{Related Work}

\paragraph{Quantization of Diffusion Models.}
Early post-training quantization methods for diffusion models focused on the
temporal non-stationarity induced by iterative denoising. Q-Diffusion addresses
timestep-dependent activation distributions through timestep-aware calibration
and split shortcut quantization, while PTQD explicitly models and corrects the
accumulation of quantization noise along the denoising trajectory
\citep{li2023qdiffusion,he2023ptqd}. As transformer-based backbones became
increasingly prominent in diffusion models, subsequent work identified
additional challenges arising from activation outliers, timestep-dependent
distributions, and heterogeneous layer sensitivity. ViDiT-Q
\citep{zhao2025viditq} introduces DiT-specific quantization schemes together
with sensitivity-aware mixed-precision allocation. DiRotQ
\citep{sharify2026dirotq} employs rotation-aware activation quantization to
separate dominant activation directions from the low-bit subspace, whereas
OrbitQuant \citep{lee2026orbitquant} quantizes weights and activations in a
normalized rotated representation using a shared non-uniform codebook.
\vspace{-0.5em}
\paragraph{Low-Rank-Assisted Quantization.}
Low-rank correction provides a complementary strategy for aggressive quantization by reserving a small auxiliary branch for structured error that is difficult to represent with low-bit codes. SVDQuant establishes this paradigm for diffusion models by concentrating weight outliers into a high-precision SVD branch and quantizing the remaining residual weights and activations to 4 bits \citep{li2025svdquant}. Recent low-rank PTQ methods optimize different parts of this calibration problem. In language models, LRC jointly optimizes quantized weights with full-precision low-rank corrections acting on unquantized activations \citep{scetbon2024lrc}. LoRaQ directly optimizes a low-rank approximation to quantization error, allowing the auxiliary branch to operate below 16-bit precision \citep{bouquet2026loraq}. 
GPTQ-intrinsic LoRA incorporates low-rank correction into an
augmented-Hessian quantization pass \citep{zhang2026gptqintrinsic}.

\vspace{-0.5em}
\paragraph{Activation-aware calibration PTQ.}
A separate line of work exploits calibration statistics to reshape the quantization problem. GPTQ uses approximate second-order information to minimize weight-quantization error under input-dependent curvature \citep{frantar2022gptq}, while AWQ identifies salient weights using activation statistics and protects them through channel-wise scaling \citep{lin2024awq}. SmoothQuant migrates activation outliers into weights through an equivalent diagonal rescaling \citep{xiao2023smoothquant}. Rotation-based methods such as QuaRot and SpinQuant instead exploit orthogonal invariances to suppress activation outliers before low-bit quantization \citep{ashkboos2024quarot,liu2025spinquant}. 

Our work shares the use of low-rank compensation and activation-aware second-order information, but focuses on their coupled W4A4 formulation and studies this coupling under small auxiliary rank budgets for efficient PTQ in diffusion transformers.

\section{Method}
\label{sec:method}
\subsection{Problem Modeling}
\label{sec:problem_setup}


For a linear layer $Y=XW$ with calibration input $X\in\mathbb{R}^{m\times c}$ and weight matrix $W\in\mathbb{R}^{c\times n}$, $r\ll\min(c,n)$ denotes the protected rank budget. We apply an equivalent diagonal reparameterization
\begin{equation}
    \widetilde X = XD^{-1},
    \qquad
    \widetilde W = DW,
    \label{eq:smoothing}
\end{equation}
where $D=\operatorname{diag}(d)$ with $d_i>0$ is the diagonal smoothing matrix. This transform redistributes channel ranges between activations and weights while preserving $XW=\widetilde X\widetilde W$, where $\widetilde X, \widetilde W$ are the smoothed activations and weights.

\paragraph{Low-rank-assisted operator.}
We adopt a low-rank-assisted operator with a rank-$r$ high-precision branch, keeping $\widetilde X L_1L_2$ separate from the low-bit path. The resulting operator is
\begin{equation}
    \widehat Y
    =
    \widetilde X L_1L_2
    +
    Q_A(\widetilde X)\widehat R,
    \label{eq:deploy}
\end{equation}
where $L_1\in\mathbb{R}^{c\times r}$ and $L_2\in\mathbb{R}^{r\times n}$ are the input- and output-side low-rank factors, respectively, and $\widehat R$ denotes the quantized residual. \rev{We then derive the reconstruction objective given by this low-rank-assisted operator.}

\paragraph{Reconstruction objective.}
Defining remaining weight error $C = \widetilde W-L_1L_2-\widehat R$ and activation error $E_A = Q_A(\widetilde X)-\widetilde X$, Equation~\eqref{eq:deploy} gives exact layer-output error
\begin{equation}
    Y-\widehat Y = \widetilde X C-E_A\widehat R.
    \label{eq:exact_error}
\end{equation}
\rev{Let $\widetilde H = \frac{1}{m}\widetilde X^\top\widetilde X$ and $\Sigma_A = \frac{1}{m}E_A^\top E_A$, with $\|B\|_M^2 \triangleq \operatorname{tr}(B^\top M B)$ for PSD matrix $M$. Expanding Equation~\eqref{eq:exact_error} yields
\begin{equation}
\label{eq:exact_squared_error}
\frac{1}{m}\|Y-\widehat Y\|_F^2 = \|C\|_{\widetilde H}^2 + \|\widehat R\|_{\Sigma_A}^2 - \frac{2}{m}\operatorname{tr}\left(C^\top \widetilde X^\top E_A \widehat R\right).
\end{equation}
Under standard conditional zero-mean uniform rounding assumptions, the cross term vanishes in expectation ($\mathbb E[\operatorname{tr}(C^\top \widetilde X^\top E_A \widehat R)] = 0$), yielding $\frac{1}{m}\|Y-\widehat Y\|_F^2 \approx \|C\|_{\widetilde H}^2 + \|\widehat R\|_{\Sigma_A}^2$.}
For a $b_A$-bit symmetric quantizer with $\kappa_A=2^{b_A-1}-1$ and activation group $\mathcal G(i)$ for channel $i$, we model activation-noise covariance via the block-constant diagonal surrogate:
\begin{equation}
    \overline\Sigma_A=\operatorname{Diag}(s_A),
    \qquad
    (s_A)_i = \frac{1}{12\kappa_A^2} \mathbb E_t\!\left[\max_{j\in\mathcal G(i)}\widetilde X_{tj}^{2}\right].
    \label{eq:sigma_model}
\end{equation}
Introducing weighting parameter $\lambda_A\geq0$, the joint calibration objective is
\begin{equation}
    \mathcal L(L_1,L_2,\widehat R) = \frac{1}{2}\left\|\widetilde W-L_1L_2-\widehat R\right\|_{\widetilde H}^{2} + \frac{1}{2}\lambda_A\left\|\widehat R\right\|_{\overline\Sigma_A}^{2}.
    \label{eq:joint_obj}
\end{equation}
\rev{Equation~\eqref{eq:joint_obj} unifies low-rank compensation and activation-noise suppression under a shared calibration framework where the first term minimizes weight error, while the second suppresses activation-error amplification through $\widehat R$. We use Equation~\eqref{eq:joint_obj} as the common objective underlying the derivations in Section~\ref{sec:hsvdquant}; \(\lambda_A=0\) gives its weight-centric specialization, while \(\lambda_A>0\) enables activation-aware code optimization when activation-quantization error is propagated during calibration.}

\subsection{H-SVDQuant}
\label{sec:hsvdquant}
We now derive tractable updates for the coupled objective in Equation~\eqref{eq:joint_obj}. The derivation proceeds by analytically eliminating the refittable factor \(L_2\), reducing residual-code optimization to a second-order metric-and-target problem, while deriving calibration-aware updates for the protected subspace and smoothing transform.

\paragraph{Hessian deflation.} For fixed $D$, $L_1$, and $\widehat R$, minimizing over $L_2$ yields the closed-form weighted least-squares solution:
\begin{equation}
    L_2^\star = \left( L_1^\top\widetilde H L_1 \right)^\dagger L_1^\top\widetilde H \left( \widetilde W-\widehat R \right),
    \label{eq:l2_opt}
\end{equation}
where $(\cdot)^\dagger$ denotes the Moore--Penrose pseudoinverse.

\noindent
\textbf{Proposition 1 (Hessian deflation).}
Substituting $L_2^\star$ into Equation~\eqref{eq:joint_obj} replaces $\widetilde H$ with the deflated Hessian
\begin{equation}
    \widetilde H_\perp = \widetilde H - \widetilde H L_1 \left( L_1^\top\widetilde H L_1 \right)^\dagger L_1^\top\widetilde H, \qquad \text{satisfying} \quad \widetilde H_\perp L_1 = 0.
    \label{eq:deflation}
\end{equation}
\rev{The derivation of $L_2^\star$ and proof are provided in Appendix~\ref{app:deflation}.}
\rev{The identity $\widetilde H_\perp L_1=0$ shows that errors compensable by the protected low-rank component are eliminated from the residual objective; residual quantization thus targets only the remaining reconstruction error.}

\paragraph{Residual-code optimization.}
After eliminating $L_2$, the code-dependent objective becomes
\begin{equation}
    \mathcal L_{\mathrm{code}}
    =
    \frac{1}{2}
    \left\|
        \widetilde W-\widehat R
    \right\|_{\widetilde H_\perp}^{2}
    +
    \frac{1}{2}
    \lambda_A
    \left\|
        \widehat R
    \right\|_{\overline\Sigma_A}^{2}
    =
    \frac{1}{2}
    \left\|
        \widehat R-T_Q
    \right\|_{M_Q}^{2}
    +
    \mathrm{const},
    \label{eq:code_obj}
\end{equation}
where 
\begin{equation}
    M_Q = \widetilde H_\perp + \lambda_A\overline\Sigma_A, \qquad T_Q = M_Q^\dagger \widetilde H_\perp \widetilde W. 
    \label{eq:mq_tq}
\end{equation} 
We use Equation~\eqref{eq:sigma_model} as the activation-noise surrogate: $M_Q$ reshapes the metric geometry, while $T_Q$ adjusts the effective approximation target. 

\paragraph{Low-rank initialization.}
Rather than initializing the protected branch with ordinary truncated SVD, we solve a calibration-weighted rank-$r$ approximation $\min_{\operatorname{rank}(L)\leq r} \frac{1}{2}\|\widetilde H_{\mathrm{reg}}^{1/2} (\widetilde W-L)\|_F^2$, whose exact solution is , where $\widetilde H_{\mathrm{reg}}=\widetilde H+\tau I$ with $\tau>0$ is a positive-definite regularization of the calibration Gram matrix. The exact solution to this regularized approximation problem is
\begin{equation}
    L^\star
    =
    \widetilde H_{\mathrm{reg}}^{-1/2}
    \left[
        \widetilde H_{\mathrm{reg}}^{1/2}
        \widetilde W
    \right]_r,
    \label{eq:weighted_init_solution}
\end{equation}
where $[\cdot]_r$ denotes rank-$r$ truncated SVD. We initialize $L_1$ with the resulting input-side span under an $\widetilde H_{\mathrm{reg}}$-orthonormal gauge, preserving $L_1L_2$ while improving numerical conditioning. Details are in Appendix~\ref{sec:protected-subspace}. 

\paragraph{Diagonal smoothing.}
The diagonal reparameterization in Equation~\eqref{eq:smoothing} redistributes channel ranges without changing the full-precision linear map. Given residual weight $P$ and diagonal curvature $h_i$, a closed-form \(\ell_p\) relaxation of the grouped geometric-program objective yields
\begin{equation}
    d_i \propto
    \left(
        {h_i} \Big/ {\textstyle\sum_j |P_{ij}|^p}
    \right)^{\!1/(p+2)}.
    \label{eq:d_closed_form}
\end{equation}
We normalize the resulting scales by their geometric mean and optionally refine them in log space using the grouped objective; Appendix~\ref{sec:smoothing-optimization} gives the exact GP formulation and the derivation of the closed-form relaxation.

\subsection{Implemented Algorithm}
\label{sec:calibration}

Based on the core derivations and formulas, we compose the conditional updates into the overall calibration algorithm and describe its numerical treatment and deployed operator.

\paragraph{Algorithm workflow.}
As summarized in Appendix Algorithm~\ref{alg:hsvdquant_detailed}, the unsmoothed calibration matrix $H_0=X^\top X/m$ is accumulated once. At each outer iteration, we update smoothing matrix $D$, obtain $\widetilde H=D^{-1}H_0D^{-1}$, initialize $L_1$, construct the deflated Hessian $\widetilde H_\perp$, quantize residual $\widehat R$, and refit $L_2$ in closed form. With cached activations, candidate states are compared using empirical W4A4 output MSE with the dynamic quantizer.
For activation-aware W4A4 calibration, we use a GPTQ-style solver with $M_Q$ and $T_Q$ replacing the conventional curvature matrix and weight target. When $\lambda_A=0$, the solver reduces to weight-only residual quantization under $\widetilde H_\perp$, followed by the same closed-form refit of $L_2$. The two settings optimize the same underlying low-rank-assisted operator but differ in whether activation-noise amplification is included in residual-code selection. 

\paragraph{Numerical stability and damping.}
The identities in Eqs.~\eqref{eq:deflation}--\eqref{eq:code_obj} describe the undamped formulation.
Since $\widetilde H_\perp$ and $M_Q$ can be singular, practical solves use small diagonal damping ($M_Q+\epsilon I$) before Cholesky factorization or inversion when computing $T_Q$, which slightly perturbs $\widetilde H_\perp L_1=0$ while ensuring stability.

\FloatBarrier

\section{Experimental Setup}
\label{sec:experiments}

\paragraph{Models.}
\rev{We benchmark \method{} across: PixArt-$\Sigma$ (0.6B)~\citep{chen2024pixartsigma}, SANA-1.6B~\citep{xie2024sana}, FLUX.1-dev (12B)~\citep{blackforestlabs2024flux}, and the Qwen-Image (20B)~\citep{qwenimage2024} (50-step base model). Accelerated few-step variants (the 4-step FLUX.1-schnell and 4-step Qwen-Image Lightning) are evaluated in Appendix Table~\ref{tab:semantic_complete}, and visual generation comparisons for both base and Lightning models are detailed in Appendix~\ref{app:qwen_qualitative}. Additionally, we evaluate Qwen3-8B~\citep{yang2025qwen3technicalreport} as a controlled testbed for rank efficiency in large language models.}

\paragraph{Implementation details.}
All diffusion backbones operate in the W4A4 post-training quantization regime with group size 64 and an auxiliary protected rank $r=4$ (FP16/BF16). 
Activations are quantized using per-token, per-group 4-bit quantization. 
In our primary DiT experiments,
we use reference activations across transformer blocks with \(\lambda_A=0\) (except for PixArt-$\Sigma$ at \(\lambda_A=0.05\) due to its different activation geometry) in Table ~\ref{tab:dit_main}. We sequentially replay activations within each block as module groups are quantized and replaced. These can be found in our Algorithm~\ref{alg:hsvdquant_detailed}. We analyze the effect and strength of \(\lambda_A\) in Section~\ref{sec:pixart_activation_weight}, and compare different strategies for propagating quantized activations and accumulating quantization error in Appendix~\ref{app:block}.

\paragraph{Datasets.}
Following SVDQuant~\citep{li2025svdquant}, we evaluate the generation fidelity and quality of quantized diffusion models on the category-balanced
\textbf{MJHQ-30K}~\citep{li2024playground} and the diverse
\textbf{sDCI}~\citep{urbanek2023picture} benchmarks.
For each diffusion backbone, evaluation is performed on fixed
\textbf{MJHQ-5K} and \textbf{sDCI-1K} subsets.
Each prompt is paired with the output of the corresponding native high-precision teacher model using identical random seeds, initial latents, image resolution ($1024\times1024$), and scheduler steps.
For diffusion-model calibration, we use the same set of 128 prompts released in the official SVDQuant repository. For Qwen3-8B calibration, we sample 128 sequences of 512 tokens each from the WikiText-2 training split. We report WikiText-2~\citep{merity2016wikitext} perplexity together with zero-shot accuracy on five standard benchmarks:
MMLU~\citep{hendrycks2020mmlu},
ARC-Easy and ARC-Challenge~\citep{clark2018arc},
HellaSwag~\citep{zellers2019hellaswag}, and
PIQA~\citep{bisk2020piqa}.

\paragraph{Baselines.}
We compare \method{} against representative post-training quantization (PTQ) baselines.
\textbf{SVDQuant}~\citep{li2025svdquant}:
we use rank-4 SVDQuant as the primary matched-capacity baseline and additionally include rank-32 SVDQuant, which uses an $8\times$ larger protected-rank budget, as a higher-capacity reference.
\textbf{DiRotQ}~\citep{sharify2026dirotq}:
we evaluate its W4A4 variant while matching the fraction of high-precision values to that of our method.
\textbf{OrbitQuant}~\citep{lee2026orbitquant} and
\textbf{ViDiT-Q}~\citep{zhao2025viditq}:
we adapt both methods to the same W4A4 setting with group size 64 and per-token activation quantization for a controlled comparison.

\paragraph{Metrics.}
Following previous works~\citep{li2025svdquant}, we assess generation fidelity and quality using four standard metrics:
\textbf{Peak Signal-to-Noise Ratio (PSNR, $\uparrow$)} measures pixel-level numerical reconstruction fidelity with respect to the 16-bit teacher model;
\textbf{Learned Perceptual Image Patch Similarity} (\textbf{LPIPS}, $\downarrow$; ~\citep{lpips}) 
evaluates perceptual similarity against prompt-paired teacher outputs;
\textbf{CLIP Score (CLIP, $\uparrow$)}~\citep{clip} measures text--image semantic alignment with the input prompts; and
\textbf{ImageReward (IR, $\uparrow$)}~\citep{imagereward} approximates human visual preference for generated images.
For language modeling on Qwen3-8B, we report WikiText-2 \textbf{Perplexity (PPL, $\downarrow$)} and zero-shot accuracy. 
For the PixArt-$\Sigma$ ablation in Table~\ref{tab:lambda_pixart}, we additionally report Structural Similarity (SSIM, $\uparrow$)~\citep{wang2004image}.
\begin{table}[!t]
\centering
\fontsize{8}{9.5}\selectfont
\setlength{\tabcolsep}{2pt}
\renewcommand{\arraystretch}{1.05}
\caption{\textbf{W4A4 fidelity and generation quality across modern diffusion transformer backbones.}
Bold and underline denote the best and second-best results, respectively.
All evaluated methods use 4-bit weights and 4-bit activations (W4A4) with group-64.}
\label{tab:dit_main}
\begin{tabular*}{\linewidth}{@{\extracolsep{\fill}}lrrrrrrrr@{}}
\toprule
& \multicolumn{4}{c}{MJHQ} & \multicolumn{4}{c}{sDCI} \\
\cmidrule(lr){2-5}\cmidrule(lr){6-9}
Method & PSNR$\uparrow$ & LPIPS$\downarrow$ & CLIP$\uparrow$ & IR$\uparrow$ & PSNR$\uparrow$ & LPIPS$\downarrow$ & CLIP$\uparrow$ & IR$\uparrow$ \\
\midrule
\multicolumn{9}{@{}l}{\textbf{PixArt-$\Sigma$}} \\[1pt]
\textbf{H-SVDQuant $r=4$} & \textbf{16.47} & \underline{0.386} & \underline{25.995} & 0.8852 & 15.36 & 0.422 & 26.203 & \underline{0.9816} \\
SVDQuant $r=4$ & 14.94 & 0.478 & 25.832 & 0.7489 & 13.75 & 0.521 & 25.604 & 0.8514 \\
SVDQuant $r=32$ & 16.27 & \textbf{0.385} & 25.962 & \underline{0.8950} & \textbf{15.45} & \textbf{0.413} & 25.718 & 0.9002 \\
DiRotQ & \underline{16.36} & 0.389 & \textbf{26.047} & \textbf{0.9165} & \underline{15.39} & \underline{0.416} & \underline{26.239} & \textbf{1.0417} \\
OrbitQuant & 14.88 & 0.488 & 25.703 & 0.7711 & 14.00 & 0.528 & \textbf{26.441} & 0.8695 \\
\midrule
\multicolumn{9}{@{}l}{\textbf{SANA-1.6B}} \\[1pt]
\textbf{H-SVDQuant $r=4$} & \textbf{20.25} & \textbf{0.161} & 25.996 & 1.1053 & \textbf{18.25} & \textbf{0.185} & 26.259 & 1.0724 \\
SVDQuant $r=4$ & 19.64 & 0.176 & 25.956 & \textbf{1.1213} & 17.13 & 0.216 & \underline{26.336} & 1.0750 \\
SVDQuant $r=32$ & \underline{20.24} & \underline{0.162} & 25.931 & 1.0994 & \underline{17.69} & \underline{0.199} & 26.330 & \textbf{1.1102} \\
DiRotQ & 19.83 & 0.173 & \underline{26.051} & \underline{1.1064} & 17.61 & 0.209 & 26.274 & \underline{1.1059} \\
OrbitQuant & 14.48 & 0.415 & \textbf{26.124} & 1.0465 & 12.10 & 0.495 & \textbf{26.701} & 1.0809 \\
ViDiT-Q & 13.38 & 0.444 & 25.721 & 0.9871 & 11.44 & 0.483 & 25.783 & 0.8971 \\
\midrule
\multicolumn{9}{@{}l}{\textbf{FLUX.1-dev}} \\[1pt]
\textbf{H-SVDQuant $r=4$} & \underline{20.58} & \underline{0.227} & 23.078 & \textbf{0.9917} & \underline{18.93} & \underline{0.250} & 24.586 & 1.0205 \\
SVDQuant $r=4$ & 18.77 & 0.303 & \underline{23.156} & 0.9771 & 17.61 & 0.319 & \underline{24.675} & \textbf{1.0630} \\
SVDQuant $r=32$ & 19.19 & 0.282 & 23.107 & 0.9782 & 18.42 & 0.274 & \textbf{24.730} & \underline{1.0594} \\
DiRotQ & \textbf{20.95} & \textbf{0.224} & \textbf{23.534} & \underline{0.9864} & \textbf{19.92} & \textbf{0.218} & 24.471 & 1.0181 \\
OrbitQuant & 19.44 & 0.279 & 22.877 & 0.9637 & 18.25 & 0.284 & 24.496 & 1.0476 \\
ViDiT-Q & 15.68 & 0.518 & 22.122 & 0.6232 & 14.49 & 0.539 & 24.140 & 0.7236 \\
\midrule
\multicolumn{9}{@{}l}{\textbf{Qwen-Image (50 steps)}} \\[1pt]
\textbf{H-SVDQuant $r=4$} & 19.23 & \underline{0.236} & 29.669 & \textbf{1.5775} & \textbf{18.44} & \textbf{0.254} & \textbf{25.817} & 1.2027 \\
SVDQuant $r=4$ & 18.78 & 0.256 & 29.714 & 1.5540 & 17.70 & 0.279 & \underline{25.755} & \textbf{1.2191} \\
SVDQuant $r=32$ & \underline{19.37} & 0.243 & \underline{29.734} & \underline{1.5665} & 18.03 & 0.272 & 25.681 & \underline{1.2095} \\
DiRotQ & \textbf{19.63} & \textbf{0.218} & \textbf{29.769} & 1.5404 & \underline{18.39} & \underline{0.257} & 25.642 & 1.2080 \\
\bottomrule
\end{tabular*}
\end{table}

\begin{figure}[!p]
\centering
\includegraphics[width=0.95\linewidth,height=0.83\textheight,keepaspectratio]{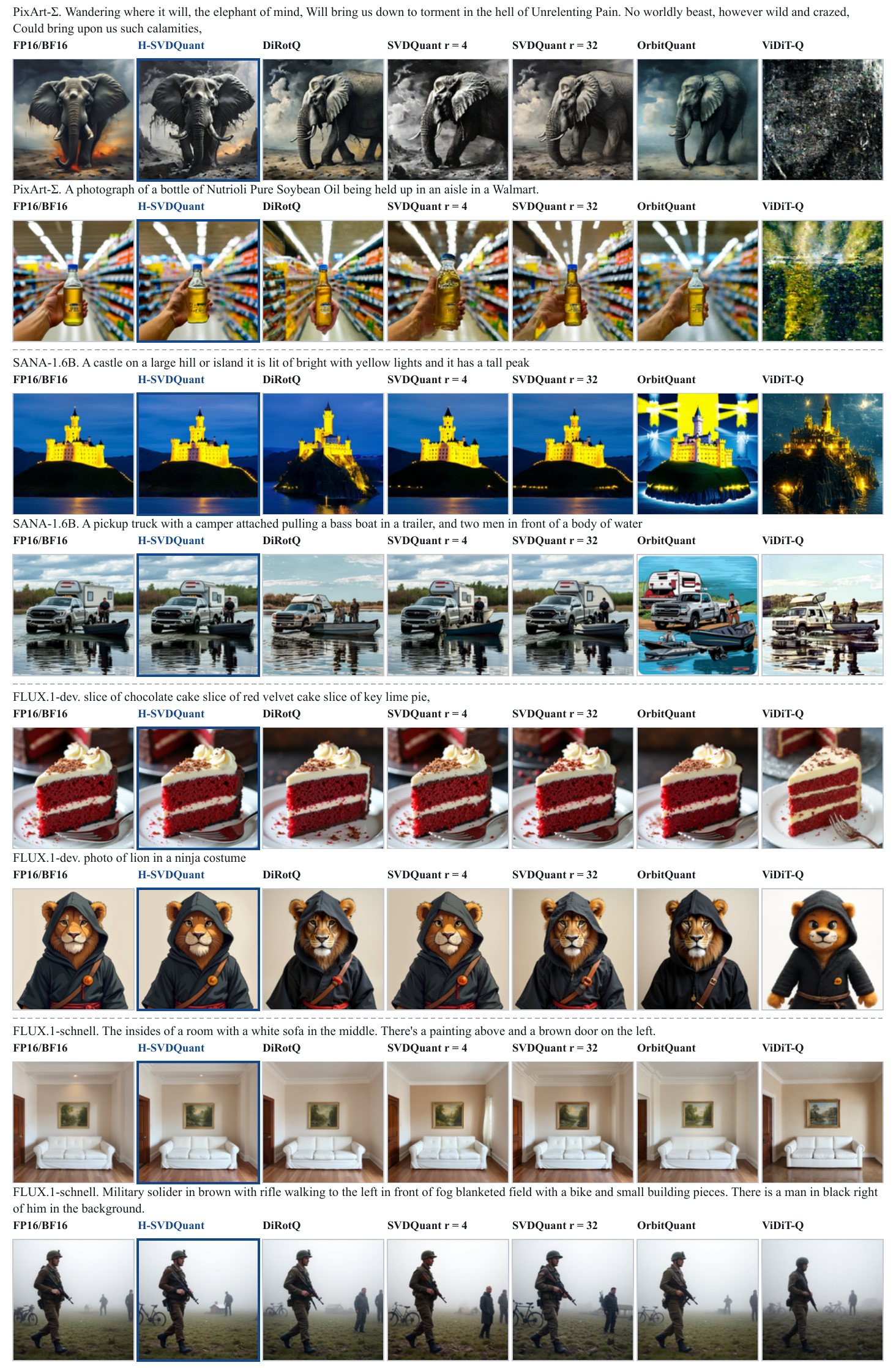}
\caption{\textbf{W4A4 reconstructions across diffusion transformers.} Each row compares dense FP16/BF16 against quantized methods under identical prompt and seed. Blue boxes highlight \method{} ($r=4$). Across all backbones, \method{} preserves textures, fine structures, and prompt alignment faithful to the 16-bit reference.}
\label{fig:qual_full_methods}
\end{figure}

\section{Main Results on Diffusion Transformers}
\label{sec:dit_results}

We report quantitative W4A4 results in Table~\ref{tab:dit_main} across DiT backbones, and show corresponding qualitative visual comparisons in Figure~\ref{fig:qual_full_methods}. Note that PSNR and LPIPS measure teacher fidelity, while CLIP and IR measure standalone generation quality.
They measure different properties; thus the method with the best PSNR need not have the highest CLIP or ImageReward score.

\paragraph{Quantitative Comparison.}
As shown in Table~\ref{tab:dit_main}, under W4A4 \method{} consistently improves teacher fidelity over SVDQuant on both MJHQ and sDCI, achieving higher PSNR and lower LPIPS on every evaluated backbone.  
On PixArt-\(\Sigma\), \method{} improves over rank-4 SVDQuant by \(+1.53\) dB PSNR and \(-0.092\) LPIPS on MJHQ, and by \(+1.61\) dB and \(-0.099\) LPIPS on sDCI. Its teacher-fidelity results are close to rank-32 SVDQuant and DiRotQ, while CLIP and ImageReward vary across methods.  
On SANA-1.6B, \method{} reaches 20.25/0.161 PSNR/LPIPS on MJHQ and 18.25/0.185 on sDCI, improving substantially over rank-4 SVDQuant and matching or exceeding rank-32 SVDQuant in teacher-fidelity metrics.  
On FLUX.1-dev, \method{} improves rank-4 SVDQuant by \(+1.81\) dB PSNR and \(-0.076\) LPIPS on MJHQ, and by \(+1.32\) dB and \(-0.069\) LPIPS on sDCI. It also exceeds rank-32 SVDQuant in PSNR and LPIPS on both subsets. 
On Qwen-Image, \method{} improves teacher fidelity over rank-4 SVDQuant on both subsets. Relative to rank-32 SVDQuant, the comparison is mixed on MJHQ—\method{} obtains lower LPIPS but slightly lower PSNR—while it improves both PSNR and LPIPS on sDCI. Extended quantitative comparisons on accelerated variants are provided in Appendix Table~\ref{tab:semantic_complete}.


\paragraph{\rev{Qualitative Comparison.}}
\rev{Figure~\ref{fig:qual_full_methods} provides visual comparisons across representative prompts under identical random seeds.
OrbitQuant and ViDiT-Q struggle under 4-bit activations, exhibiting pervasive noise, structural distortion, and contrast loss (e.g., distorted geometry on PixArt-$\Sigma$ and noisy surfaces on SANA-1.6B and FLUX.1-dev).
While DiRotQ and rank-32 SVDQuant achieve better visual quality, they lose fidelity in a few cases (e.g., different elephant direction and hill space, over-detailed lion).
In contrast, \method{} consistently preserves sharp object boundaries and subtle textures faithful to the 16-bit reference teacher. Extended visual comparisons on both base and Lightning Qwen-Image models are provided in Appendix~\ref{app:qwen_qualitative}.}

\paragraph{Rank Efficiency and Calibration Cost.}
The rank advantage is clearest on SANA and the FLUX backbones. 
Besides the mean values, we also provide the paired distributions in Figure~\ref{fig:intro_method_pair}(b) and Appendix Figure~\ref{fig:paired_medium} for a clearer comparison.
As shown in Table~\ref{tab:ptq_main}, \method{} reduces offline PTQ calibration cost in the recorded base runs. On SANA-1.6B, the measured wall time decreases from 30.0 to 4.8 minutes, while the FLUX.1-schnell and FLUX.1-dev base runs show 1.81$\times$ and 1.85$\times$ speedups, respectively.

\begin{table}[!t]
\centering\small
\caption{\textbf{Recorded rank-4 PTQ cost on NVIDIA H20.}
Time is each checkpoint's complete launcher wall interval; peak memory is the recorded CUDA allocation.}
\label{tab:ptq_main}
\begin{tabular}{llrrr}
\toprule
Model & Method & Minutes & Speedup vs. SVDQuant r=4 & Peak GiB\\
\midrule
SANA-1.6B & H-SVDQuant & 4.8 & $6.25\times$ & 9.9\\
& SVDQuant & 30.0 & -- & 11.7\\
FLUX-schnell & H-SVDQuant & 282.4 & $1.81\times$ & 52.0\\
& SVDQuant & 511.1 & -- & 50.7\\
FLUX-dev & H-SVDQuant & 285.6 & $1.85\times$ & 52.0\\
& SVDQuant & 527.4 & -- & 51.3\\
\bottomrule
\end{tabular}
\end{table}

\begin{table}[!t]
\centering
\caption{\textbf{Qwen3-8B zero-shot evaluation (W4A4, group-128).} Mean denotes the unweighted average across all five downstream tasks. SVDQuant refers to the local language-model adaptation; FP16 serves as dense reference.}
\label{tab:qwen_eval}
\small
\setlength{\tabcolsep}{4.5pt}
\begin{tabular}{lcccccc}
\toprule
Method & MMLU & ARC-C & ARC-E & HellaSwag & PIQA & Mean \\
\midrule
FP16 & 72.95 & 56.40 & 83.54 & 74.97 & 76.88 & 72.95 \\
SVDQuant adaptation $r=4$ & 61.34 & 45.31 & 73.02 & 67.83 & 71.76 & 63.85 \\
SVDQuant adaptation $r=32$ & 61.50 & 48.89 & 74.83 & 67.84 & 72.20 & 65.05 \\
\method{} $r=4$ & \textbf{68.17} & \textbf{49.83} & \textbf{78.54} & \textbf{70.30} & \textbf{73.34} & \textbf{68.04} \\
\bottomrule
\end{tabular}
\end{table}

\section{Analysis and Ablation Studies}
\label{sec:analysis}
In this section, we conduct in-depth diagnostics and ablation studies using PixArt-$\Sigma$ (for vision generation) and Qwen3-8B (for language modeling). Unless stated, our optional cached iteration of $\mD$ is 20 and the outer iteration is set to be $2$. In Qwen analysis, we disable the optional cached optimization and set outer iteration as 1 in Table \ref{tab:qwen_diagnostics}(a), \ref{tab:lambda}(b), \ref{tab:deflation_ablation} and set outer iteration as 2 in Table \ref{tab:qwen_d_ablation}.
We explore the effect of rank $r$ and $\lambda_A$, and demonstrate the  advantages of our three optimization components: $\mH_\perp$, $\mL_1$, $\mD$.
\paragraph{{Rank Sweep Ablation.}}
\label{sec:qwen_rank_efficiency}
We first examine the protected low-rank budget $r$. As detailed in Table~\ref{tab:qwen_diagnostics}(a), rank-0 \method{} already achieves 10.72 perplexity, substantially outperforming rank-32 SVDQuant (11.21). Introducing a small protected branch provides an additional improvement, reducing perplexity to 10.44 (10.438). As shown in Table~\ref{tab:qwen_diagnostics}(b), the rank-4 configuration also exhibits substantially lower local and propagated activation error than the SVDQuant baselines. Together, these results show that \method{} remains effective under very small protected-rank budgets, characterizing the complete \method{} recipe rather than attributing the gain to a single component.
\begin{table}[!t]
\centering
\caption{
\textbf{Rank efficiency and activation-error propagation
on Qwen3-8B} (W4A4, group-128, $\lambda_A=0.25$).
(a) WikiText-2 perplexity across different protected ranks.
(b) Relative activation MSE averaged over 36 transformer blocks
using eight 512-token sequences (measured against 16-bit reference activations).
Local MSE measures per-block quantization error under reference
inputs, while propagated MSE additionally captures errors
accumulated from preceding quantized blocks.
}
\label{tab:qwen_diagnostics}
\begin{subtable}[t]{0.40\linewidth}
\centering\small
\setlength{\tabcolsep}{3pt}
\caption{Perplexity across protected ranks}
\label{tab:qwen_ppl}
\begin{tabular}{lrr}
\toprule
Method & Rank & PPL $\downarrow$ \\
\midrule
SVDQuant adapt. & 4 & 11.93 \\
SVDQuant adapt. & 32 & 11.21 \\
\method{} & 0 & 10.72 \\
\method{} & 2 & 10.48 \\
\method{} & 4 & \textbf{10.44} \\
\bottomrule
\end{tabular}
\end{subtable}
\hfill
\begin{subtable}[t]{0.58\linewidth}
\centering\small
\setlength{\tabcolsep}{3pt}
\caption{Local vs. propagated activation error}
\label{tab:act_probe}
\begin{tabular}{lrr}
\toprule
Method & Local MSE $\downarrow$ & Prop. MSE $\downarrow$ \\
\midrule
SVDQuant adapt. $r=4$ & 0.00215 & 0.02068 \\
SVDQuant adapt. $r=32$ & 0.00199 & 0.01715 \\
\method{} $r=4$ & \textbf{0.00150} & \textbf{0.01194} \\
\midrule
\textit{Reduction vs.\ $r=4$} & \textbf{$-$30.2\%} & \textbf{$-$42.3\%} \\
\textit{Reduction vs.\ $r=32$} & \textbf{$-$24.6\%} & \textbf{$-$30.4\%} \\
\bottomrule
\end{tabular}
\end{subtable}
\end{table}

\paragraph{{Activation-Noise Sensitivity.}}
\label{sec:pixart_activation_weight}
We ablate the activation-noise weighting parameter in Equation~\eqref{eq:joint_obj} across both PixArt-$\Sigma$ and Qwen3-8B (Table~\ref{tab:lambda}), with $\lambda_A \in \{0, 0.05, 0.10, 0.25, 0.50\}$.
On PixArt-$\Sigma$ (Table~\ref{tab:lambda}(a)), \(\lambda_A\) has a modest effect. A small activation-noise weight improves all three metrics at once: they attain their best values at $\lambda_A = 0.05$ (15.99~dB PSNR, 0.583 SSIM, and 0.414 LPIPS), a $+0.29$~dB PSNR gain over $\lambda_A = 0$. Raising $\lambda_A$ further gives this gain back, and $\lambda_A = 0.50$ falls slightly below the unweighted baseline (15.37~dB PSNR, 0.554 SSIM, 0.472 LPIPS).
On Qwen3-8B (Table~\ref{tab:lambda}(b)), WikiText-2 perplexity remains highly stable across small weighting coefficients ($\lambda_A \le 0.25$), reaching its lowest value of 10.433 at $\lambda_A = 0.05$ and ranking second at $\lambda_A = 0.25$ (10.438).
Empirical results demonstrate that moderate activation-aware weighting ($\lambda_A \in [0.05, 0.25]$) provides the most effective balance, whereas excessive weighting ($\lambda_A = 0.50$) leads to performance regression. The benefit of activation-aware weighting is therefore regime- and model-dependent.

\begin{table}[!t]
\centering
\caption{\textbf{Activation-noise-weight sensitivity on PixArt-$\Sigma$ (group-64), and Qwen3-8B (group-128). (W4A4, $r=4$).}
We evaluating sensitivity to activation-noise weighting parameter $\lambda_A$ across vision generation (PixArt-$\Sigma$) and language modeling (Qwen3-8B). Bold denotes best value in each backbone. In Qwen, we disable the optional cached iteration and the outer iteration, the result of $\lambda_A=0$ matches Table~\ref{tab:deflation_ablation}.}
\label{tab:lambda}
\begin{subtable}[t]{0.52\linewidth}
\centering\small
\setlength{\tabcolsep}{4.5pt}
\caption{PixArt-$\Sigma$}
\label{tab:lambda_pixart}
\begin{tabular}{lccc}
\toprule
$\lambda_A$ & PSNR $\uparrow$ & SSIM $\uparrow$ & LPIPS $\downarrow$ \\
\midrule
0     & 15.70 & 0.557 & 0.467 \\
0.05  & \textbf{15.99} & \textbf{0.583} & \textbf{0.414} \\
0.10  & 15.97 & 0.574 & 0.441 \\
0.25  & 15.73 & 0.580 & 0.433 \\
0.50  & 15.37 & 0.554 & 0.472 \\
\bottomrule
\end{tabular}
\end{subtable}
\hfill
\begin{subtable}[t]{0.40\linewidth}
\centering\small
\setlength{\tabcolsep}{4pt}
\caption{Qwen3-8B}
\label{tab:lambda_qwen}
\begin{tabular}{lc}
\toprule
$\lambda_A$ & WikiText-2 PPL $\downarrow$ \\
\midrule
0     & 10.486 \\
0.05  & \textbf{10.433} \\
0.10  & 10.454 \\
0.25  & \underline{10.438} \\
0.50  & 10.640 \\
\bottomrule
\end{tabular}
\end{subtable}
\end{table}
\begin{table}[t]
\centering
\caption{
\textbf{Controlled ablation of protected-subspace initialization (in Equation~\eqref{eq:weighted_init_solution}.) and Hessian deflation
on Qwen3-8B (W4A4, group-128, $\lambda_A=0$)}.
We keep all other calibration settings fixed. Each row of
$\widetilde \mH$/$\widetilde \mH_\perp$ runs shares the same initialized $\mL_1$ and residual.
Lower WikiText-2 PPL is better.
}
\label{tab:deflation_ablation}
\small
\begin{tabular}{lccc}
\toprule
$\mL_1$ initialization
& Original $\widetilde \mH$
& Deflated $\widetilde \mH_\perp$\\
\midrule
Plain SVD
& 10.559
& 10.539 \\
Calibration-weighted
& 10.492
& \textbf{10.486} \\
\bottomrule
\end{tabular}
\end{table}

\paragraph{Hessian Deflation and Protected-Subspace Initialization.}
Table~\ref{tab:deflation_ablation} presents a controlled $2\times2$ ablation on Qwen3-8B ($\lambda_A=0$). We fix $\mD$, $r$, the quantizer, GPTQ settings, calibration data, and $\mL_2$ refitting, varying only the $\mL_1$ initialization (plain SVD vs.\ calibration-weighted, Equation~\eqref{eq:weighted_init_solution}) and the GPTQ metric ($\widetilde{\mH}$ vs.\ $\widetilde{\mH}_\perp$, Equation~\eqref{eq:deflation}). 
For each fixed $\mL_1$, both metrics share the same initial residual. Holding $\mL_1$ fixed, Hessian deflation reduces PPL by 0.020 (plain SVD) and 0.006 (weighted SVD). Holding the GPTQ metric fixed, weighted initialization reduces PPL by 0.067 ($\widetilde{\mH}$) and 0.053 ($\widetilde{\mH}_\perp$). 
Their combination achieves the lowest PPL (10.486, reducing baseline PPL by 0.073), supporting their complementary roles: weighted initialization selects protected directions, while deflation excludes them from residual-code optimization.

\begin{table}[t]
    \centering
    \small
    \setlength{\tabcolsep}{3.8pt}
    \caption{\textbf{Effect of our closed-form smoothing solution on Qwen3-8B (W4A4/W4A16, group-128, $\lambda_A=0.25$)}. We keep the
    \method{} pipeline fixed and replace the SVDQuant grid-searched
    $\mD$ with our closed-form solution in Equation~\eqref{eq:d_closed_form}. We disable the optional cached optimization of $\mD$ and only use its initialization.
    Avg. denotes the mean accuracy over MMLU, ARC-Challenge, ARC-Easy,
    HellaSwag, and PIQA. Lower PPL and higher accuracies are better.}
    \label{tab:qwen_d_ablation}
    \resizebox{\linewidth}{!}{%
    \begin{tabular}{llccccccc}
        \toprule
        $\mD$ construction & Quant.
        & PPL $\downarrow$
        & MMLU $\uparrow$
        & ARC-C $\uparrow$
        & ARC-E $\uparrow$
        & Hella. $\uparrow$
        & PIQA $\uparrow$
        & Avg. $\uparrow$ \\
        \midrule
        Grid-search $\mD$ (SVDQuant)
        & W4A4
        & 10.895
        & 66.59
        & \textbf{51.11}
        & \textbf{79.21}
        & 68.26
        & 74.05
        & 67.84 \\
        \textbf{Closed-form $\mD$ (Ours)}
        & W4A4
        & \textbf{10.469}
        & \textbf{67.23}
        & 50.77
        & 78.16
        & \textbf{70.77}
        & \textbf{75.24}
        & \textbf{68.44} \\
        \addlinespace[2pt]
        Grid-search $\mD$ (SVDQuant)
        & W4A16
        & 10.078
        & 70.83
        & 53.75
        & 82.03
        & \textbf{73.80}
        & 76.82
        & 71.45 \\
        \textbf{Closed-form $\mD$ (Ours)}
        & W4A16
        & \textbf{9.976}
        & \textbf{70.94}
        & \textbf{54.18}
        & \textbf{82.37}
        & 73.47
        & \textbf{77.09}
        & \textbf{71.61} \\
        \bottomrule
    \end{tabular}%
    }
\end{table}
\paragraph{Closed-Form Diagonal Smoothing.} We isolate the effect of our smoothing solver by keeping the \method{} pipeline fixed and changing only how $\mD$ is obtained. Specifically, we compare (i) SVDQuant-style grid-searched $\mD$, and (ii) our analytical solution in Equation~\eqref{eq:d_closed_form}. As shown in Table~\ref{tab:qwen_d_ablation}, the closed-form solution improves the overall result under both quantization settings: for W4A4, it reduces PPL from $10.895$ to $10.469$ and improves the average accuracy from $67.84$ to $68.44$; for W4A16, it reduces PPL from $10.078$ to $9.976$ and improves the average accuracy from $71.45$ to $71.61$. The gains are particularly clear on HellaSwag and PIQA under W4A4, with improvements of $+2.51$ and $+1.19$ points, respectively, while W4A16 also improves ARC-Challenge and ARC-Easy by $+0.43$ and $+0.34$. This controlled comparison provides empirical support for Equation~\eqref{eq:d_closed_form}: the analytically derived $\mD$ provides a better overall calibration point than SVDQuant-style grid search while eliminating the associated search overhead.

\vspace{-2mm}
\section{Conclusion}
\label{sec:conclusion}

In this paper, we presented \method{}, a unified framework for low-rank-assisted W4A4 PTQ. By jointly modeling low-rank compensation, residual quantization, and activation noise, we derive a deflated Hessian through analytical elimination of the output-side low-rank factor, together with an activation-aware extension. Our efficient calibration algorithm combines weighted low-rank initialization, GP-derived smoothing, GPTQ-style quantization, and closed-form refitting. Across diffusion transformers, rank-4 \method{} consistently improves PSNR and LPIPS over rank-4 SVDQuant, matching or outperforming rank-32 SVDQuant on SANA-1.6B and FLUX variants with an 8$\times$ smaller rank. It also reduces offline calibration time by up to 6.25$\times$. Results on Qwen3-8B further demonstrate improved perplexity and downstream accuracy over rank-32 SVDQuant, suggesting broader applicability beyond diffusion transformers.

\vspace{-2mm}
\subsubsection*{Acknowledgments}
The authors would like to thank Jian Yang from Renmin University of China for his insightful discussions and general help.

\bibliography{references}
\bibliographystyle{iclr2027_conference}
\clearpage
\appendix
\begin{center}{\Large\bfseries Appendix}\end{center}
\section{Problem Reformulation and Error Decomposition}
\label{sec:reformulation}

This section uses unnormalized Gram matrices and summed reconstruction losses;
dividing both weight and activation terms by the calibration sample count
recovers the main-text convention. The provisional-residual expansion below
motivates the smoothing surrogate. Once codes are selected, the residual matrix
is held fixed while $L_2$ is refitted, as in Eq.~\eqref{eq:l2_opt}.

Consider a linear layer
\[
Y=XW,
\qquad
X\in\mathbb{R}^{m\times c},
\qquad
W\in\mathbb{R}^{c\times n}.
\]

\subsection{Smoothing and Low-Rank Decomposition}

Let
\[
D=\operatorname{diag}(d_1,\ldots,d_c),
\qquad d_i>0,
\]
and define
\[
\widetilde X=XD^{-1},
\qquad
\widetilde W=DW.
\]
Then
\[
Y=XW=\widetilde X\widetilde W.
\]

Define
\[
H=X^\top X,
\qquad
\widetilde H
=
\widetilde X^\top\widetilde X
=
D^{-1}HD^{-1}.
\]

We decompose the smoothed weight as
\[
\boxed{
\widetilde W=L_1L_2+R,
}
\qquad
R=\widetilde W-L_1L_2,
\]
where
\[
L_1\in\mathbb{R}^{c\times r},
\qquad
L_2\in\mathbb{R}^{r\times n}.
\]

The quantized forward pass is
\[
\boxed{
\widehat Y
=
\widetilde X L_1L_2
+
Q_A(\widetilde X)Q_W(R).
}
\]

\subsection{Error Decomposition}

Define the activation and weight quantization errors as
\[
E_A
=
Q_A(\widetilde X)-\widetilde X,
\qquad
\Delta_W
=
R-Q_W(R).
\]
Hence
\[
Q_A(\widetilde X)=\widetilde X+E_A,
\qquad
Q_W(R)=R-\Delta_W.
\]

Substituting these expressions into the quantized forward pass gives
\begin{align}
\widehat Y
&=
\widetilde X L_1L_2
+
(\widetilde X+E_A)(R-\Delta_W)
\nonumber\\
&=
Y
-
\widetilde X\Delta_W
+
E_A R
-
E_A\Delta_W.
\end{align}

Therefore,
\[
Y-\widehat Y
=
\widetilde X\Delta_W
-
E_A R
+
E_A\Delta_W.
\]

Neglecting the second-order term \(E_A\Delta_W\),
\begin{equation}
\boxed{
Y-\widehat Y
\approx
\widetilde X\Delta_W-E_A R.
}
\label{eq:approx-output-error}
\end{equation}

Under the additive-noise assumption, the cross term vanishes in expectation,
yielding
\begin{equation}
\boxed{
\mathcal L
\approx
F_W+F_A,
}
\label{eq:two-channel}
\end{equation}
where
\begin{equation}
\boxed{
F_W
=
\operatorname{tr}
\left(
\Delta_W^\top
\widetilde H
\Delta_W
\right),
}
\label{eq:Fw}
\end{equation}
and
\begin{equation}
\boxed{
F_A
=
\operatorname{tr}
\left(
R^\top
\Sigma_A
R
\right),
\qquad
\Sigma_A
=
\mathbb{E}[E_A^\top E_A].
}
\label{eq:Fa}
\end{equation}

Thus,
\begin{equation}
\boxed{
\mathcal L
\approx
\operatorname{tr}
\left(
\Delta_W^\top\widetilde H\Delta_W
\right)
+
\operatorname{tr}
\left(
R^\top\Sigma_A R
\right).
}
\label{eq:reformulated-loss}
\end{equation}

\section{Elimination of \texorpdfstring{$L_2$}{L2} and Hessian Deflation}
\label{sec:hessian-deflation}\label{app:deflation}
The inverse formulas below assume a nonsingular protected Gram matrix.
For the general positive-semidefinite case, replace the inverse with its
Moore--Penrose pseudoinverse: the residual is the orthogonal projection onto
the complement of $\operatorname{range}(\widetilde H^{1/2}L_1)$.
Consequently the same minimized metric and annihilation identity hold.

Fix \(D\), \(L_1\), and the quantized residual \(Q_W(R)\).
For output column \(j\), define
\[
\widetilde w_j=\widetilde W_{:,j},
\qquad
\widehat w_j=Q_W(R)_{:,j},
\qquad
\ell_j=(L_2)_{:,j},
\]
and let
\[
u_j=\widetilde w_j-\widehat w_j.
\]

The reconstruction problem associated with column \(j\) is
\begin{equation}
\min_{\ell_j\in\mathbb{R}^r}
\frac{1}{2}
\left\|
\widetilde X
\left(
u_j-L_1\ell_j
\right)
\right\|_2^2.
\label{eq:l2-column-problem}
\end{equation}

Using
\[
\widetilde H=\widetilde X^\top\widetilde X,
\]
this is equivalently
\[
\min_{\ell_j\in\mathbb{R}^r}
\frac{1}{2}
\left(
u_j-L_1\ell_j
\right)^\top
\widetilde H
\left(
u_j-L_1\ell_j
\right).
\]

Define
\[
M=L_1^\top\widetilde H L_1.
\]

Expanding the objective gives
\[
f(\ell_j)
=
\frac{1}{2}
u_j^\top\widetilde H u_j
-
\ell_j^\top L_1^\top\widetilde H u_j
+
\frac{1}{2}
\ell_j^\top M\ell_j.
\]

The first-order optimality condition is
\[
-L_1^\top\widetilde H u_j
+
M\ell_j
=
0,
\]
hence
\begin{equation}
\boxed{
\ell_j^\star
=
M^{-1}
L_1^\top\widetilde H u_j.
}
\label{eq:l2-opt}
\end{equation}

Substituting \(\ell_j^\star\) back into the objective yields
\begin{align}
f(\ell_j^\star)
&=
\frac{1}{2}
u_j^\top\widetilde H u_j
-
\frac{1}{2}
u_j^\top
\widetilde H L_1
M^{-1}
L_1^\top\widetilde H
u_j
\nonumber\\
&=
\frac{1}{2}
u_j^\top
\left[
\widetilde H
-
\widetilde H L_1
M^{-1}
L_1^\top\widetilde H
\right]
u_j.
\end{align}

This motivates the deflated Hessian
\begin{equation}
\boxed{
\widetilde H_\perp
=
\widetilde H
-
\widetilde H L_1
\left(
L_1^\top\widetilde H L_1
\right)^{-1}
L_1^\top\widetilde H.
}
\label{eq:deflated-hessian}
\end{equation}

Hence,
\begin{equation}
\boxed{
\min_{\ell_j}
\frac{1}{2}
\left(
u_j-L_1\ell_j
\right)^\top
\widetilde H
\left(
u_j-L_1\ell_j
\right)
=
\frac{1}{2}
u_j^\top
\widetilde H_\perp
u_j.
}
\label{eq:deflated-objective}
\end{equation}

Moreover,
\[
\widetilde H_\perp L_1=0,
\]
because
\begin{align}
\widetilde H_\perp L_1
&=
\widetilde H L_1
-
\widetilde H L_1
\left(
L_1^\top\widetilde H L_1
\right)^{-1}
L_1^\top\widetilde H L_1
\nonumber\\
&=0.
\end{align}

Thus, the rank-\(r\) subspace spanned by \(L_1\) lies in the null space of
the residual quantization metric. Quantization errors along these directions
can be absorbed by the high-precision low-rank branch.

Under the \(\widetilde H\)-orthogonal gauge
\[
L_1^\top\widetilde H L_1=I,
\]
the expressions simplify to
\[
\boxed{
\ell_j^\star
=
L_1^\top\widetilde H u_j,
}
\qquad
\boxed{
\widetilde H_\perp
=
\widetilde H
-
\widetilde H L_1L_1^\top\widetilde H.
}
\]

\section{Protected-Subspace Allocation}
\label{sec:protected-subspace}

The low-rank branch is not only a low-rank approximation of the weight matrix. Its
input-side factor also determines a rank-$r$ subspace in which quantization error can
be absorbed by the high-precision branch. Let
\[
\widetilde{X}=XD^{-1},
\qquad
\widetilde{W}=DW,
\qquad
\widetilde{H}=\widetilde{X}^{\top}\widetilde{X},
\]
and write
\[
\widetilde{W}=L_1L_2+R,
\qquad
L_1\in\mathbb{R}^{c\times r},
\qquad
L_2\in\mathbb{R}^{r\times n}.
\]
The protected subspace is
\[
\mathcal{S}:=\operatorname{range}(L_1)\subseteq\mathbb{R}^{c}.
\]

For a fixed quantized residual column $\widehat{w}_j$, eliminating the corresponding
column $\ell_j$ of $L_2$ gives
\[
\min_{\ell_j\in\mathbb{R}^{r}}
\frac{1}{2}
\left\|
\widetilde{X}
\left(
\widetilde{w}_j-L_1\ell_j-\widehat{w}_j
\right)
\right\|_2^2
=
\frac{1}{2}
\left\|
\widetilde{w}_j-\widehat{w}_j
\right\|_{\widetilde{H}_{\perp}}^2,
\]
where
\[
\widetilde{H}_{\perp}
=
\widetilde{H}
-
\widetilde{H}L_1
\left(
L_1^{\top}\widetilde{H}L_1
\right)^{-1}
L_1^{\top}\widetilde{H}.
\]
In particular,
\[
\widetilde{H}_{\perp}L_1=0.
\]
Hence directions inside $\mathcal{S}$ are removed from the effective Hessian seen by
the quantizer: errors along these directions are compensated by the high-precision
low-rank branch. The rank budget $r$ therefore determines how many directions can be
protected, and the central question is how to allocate this budget.

\subsection{Weighted Rank-\texorpdfstring{$r$}{r} Approximation}
\label{subsec:weighted-rank-r}

A natural starting point is to choose a rank-$r$ matrix
\[
L=L_1L_2
\]
that approximates $\widetilde{W}$. Ordinary truncated SVD solves
\[
\min_{\operatorname{rank}(L)\le r}
\frac{1}{2}
\left\|
\widetilde{W}-L
\right\|_F^2,
\]
which treats errors in all input directions equally. For layer reconstruction,
however, the relevant quantity is the output perturbation
\[
\widetilde{X}(\widetilde{W}-L).
\]
This leads to the weighted problem
\[
\min_{\operatorname{rank}(L)\le r}
\frac{1}{2}
\left\|
\widetilde{X}(\widetilde{W}-L)
\right\|_F^2.
\]
Using
\[
\widetilde{H}
=
\widetilde{X}^{\top}\widetilde{X},
\]
the objective can be written as
\[
\frac{1}{2}
\left\|
\widetilde{X}(\widetilde{W}-L)
\right\|_F^2
=
\frac{1}{2}
\operatorname{tr}
\left(
(\widetilde{W}-L)^{\top}
\widetilde{H}
(\widetilde{W}-L)
\right)
=
\frac{1}{2}
\left\|
\widetilde{H}^{1/2}(\widetilde{W}-L)
\right\|_F^2.
\]
Therefore, after the change of variable
\[
M=\widetilde{H}^{1/2}L,
\]
the problem becomes an ordinary best rank-$r$ approximation:
\[
\min_{\operatorname{rank}(M)\le r}
\frac{1}{2}
\left\|
\widetilde{H}^{1/2}\widetilde{W}-M
\right\|_F^2.
\]
By the Eckart--Young theorem,
\[
M^{\star}
=
\left[
\widetilde{H}^{1/2}\widetilde{W}
\right]_r,
\]
and hence
\[
\boxed{
L^{\star}
=
\widetilde{H}^{-1/2}
\left[
\widetilde{H}^{1/2}\widetilde{W}
\right]_r.
}
\]
Thus the weighted approximation protects directions that are simultaneously important
in the weight matrix and strongly excited by the calibration activations.

\subsection{Hessian Eigenspace Limit}
\label{subsec:hessian-eigenspace}

The weighted approximation above is still a low-rank approximation problem. A second
view is obtained by directly minimizing the expected quantization distortion after the
protected subspace has been removed.

For output column $j$, let the weight quantization error be
\[
\Delta w_j
=
w_j-\widehat{w}_j.
\]
Under an isotropic within-column quantization-noise model,
\[
\mathbb{E}
\left[
\Delta w_j\Delta w_j^{\top}
\right]
=
\frac{\delta_j^2}{12}I_c.
\]
After eliminating $L_2$, the expected distortion becomes
\[
\mathbb{E}
\left[
\Delta w_j^{\top}
\widetilde{H}_{\perp}
\Delta w_j
\right]
=
\frac{\delta_j^2}{12}
\operatorname{tr}
\left(
\widetilde{H}_{\perp}
\right).
\]
Summing over output columns gives
\[
F_W
\propto
\operatorname{tr}
\left(
\widetilde{H}_{\perp}
\right).
\]

Let
\[
\widetilde{H}
=
V\Lambda V^{\top},
\qquad
\Lambda
=
\operatorname{diag}
(\lambda_1,\ldots,\lambda_c),
\qquad
\lambda_1\ge\cdots\ge\lambda_c>0.
\]
The rank-$r$ protected subspace that minimizes
$\operatorname{tr}(\widetilde{H}_{\perp})$ is the dominant eigenspace
\[
\boxed{
\mathcal{S}^{\star}
=
\operatorname{span}
\{v_1,\ldots,v_r\}.
}
\]
Under the $\widetilde{H}$-orthogonal gauge
\[
L_1^{\top}\widetilde{H}L_1=I_r,
\]
one convenient representative is
\[
\boxed{
L_1^{\star}
=
V_r\Lambda_r^{-1/2},
}
\]
where $V_r=[v_1,\ldots,v_r]$ and
$\Lambda_r=\operatorname{diag}(\lambda_1,\ldots,\lambda_r)$.

This result gives a different interpretation from ordinary SVD. Plain SVD protects
directions with large weight energy, whereas Hessian deflation protects directions in
which a perturbation causes the largest layer-output distortion. When the rank budget
is small, protecting these stiff directions can be more valuable than minimizing the
unweighted residual norm.

\subsection{Beta-Family}
\label{subsec:beta-family}

The dynamic-range role of plain SVD and the sensitivity role of Hessian deflation can
be interpolated by a one-parameter family. For $\beta\ge 0$, define
\[
\boxed{
L^{(\beta)}
=
\widetilde{H}^{-\beta}
\left[
\widetilde{H}^{\beta}\widetilde{W}
\right]_r.
}
\]
If
\[
\widetilde{H}^{\beta}\widetilde{W}
=
U^{(\beta)}
\Sigma^{(\beta)}
V^{(\beta)\top},
\]
then an associated input-side factor is
\[
L_1^{(\beta)}
=
\widetilde{H}^{-\beta}
U_r^{(\beta)},
\]
followed, if desired, by rescaling to satisfy
\[
L_1^{(\beta)\top}
\widetilde{H}
L_1^{(\beta)}
=
I_r.
\]

The family includes two useful regimes, of which the main recipe uses $\beta=1/2$.

For $\beta=0$,
\[
L^{(0)}
=
[\widetilde{W}]_r,
\]
which is the ordinary truncated-SVD branch. This endpoint retains the leading weight singular components.

For $\beta=\frac12$,
\[
L^{(1/2)}
=
\widetilde{H}^{-1/2}
\left[
\widetilde{H}^{1/2}\widetilde{W}
\right]_r,
\]
which is exactly the weighted rank-$r$ minimizer of
\[
\min_{\operatorname{rank}(L)\le r}
\frac{1}{2}
\left\|
\widetilde{X}(\widetilde{W}-L)
\right\|_F^2.
\]

The weight-aware family should not be identified with the Hessian-only
trace-minimization problem above. In particular, the inverse factor
$\widetilde H^{-\beta}$ means that increasing $\beta$ does not generally
make the protected span converge to the leading Hessian eigenspace.
The reported experiments fix $\beta=1/2$ and do not use an adaptive
activation-rank split or a global rank allocator.

\section{Smoothing Optimization}
\label{sec:smoothing-optimization}

The smoothing variables must be optimized while the low-rank branch is held in
the original input coordinates.  Write
\[
L_1=DA,
\qquad
B=AL_2,
\qquad
P=W-B.
\]
When $A$ and $B$ are fixed, the smoothed residual is
\[
R=\widetilde W-L_1L_2=D(W-B)=DP.
\]
Moreover, the deflated Hessian in these coordinates satisfies
\begin{equation}
\boxed{
\widetilde H_\perp
=D^{-1}H_\perp D^{-1},
\qquad
H_\perp
=H-HA(A^\top HA)^{-1}A^\top H.
}
\label{eq:smoothing-deflated-hessian}
\end{equation}
Thus $H_\perp$ and $P$ remain fixed throughout the $D$-update.  This
coordinate choice also reflects the forward pass: the high-precision branch
evaluates $\widetilde X L_1L_2=XAL_2$ and does not itself need smoothing.

\subsection{Grouped Activation-Noise Model}
\label{subsec:grouped-activation-noise}

Partition the $c$ input channels into activation-quantization groups
$\mathcal G_A$.  For calibration token $t$ and group $G\in\mathcal G_A$,
consider a symmetric $b_A$-bit quantizer with
$\kappa_A=2^{b_A-1}-1$ and step size
\[
\delta_{A,t,G}(d)
=\frac{1}{\kappa_A}
\max_{i\in G}\frac{|X_{ti}|}{d_i}.
\]
Under the independent uniform-noise approximation, each channel in $G$ has
error variance $\delta_{A,t,G}^2/12$ at token $t$.  Consequently, the
grouped diagonal activation-noise covariance has entries
\begin{equation}
\boxed{
(\Sigma_A)_{ii}
=\sigma_{G(i)}^2(d),
\qquad
\sigma_G^2(d)
=\frac{1}{12\kappa_A^2}
\sum_{t=1}^{m}
\max_{k\in G}\frac{X_{tk}^2}{d_k^2}.
}
\label{eq:grouped-activation-covariance}
\end{equation}
Using $R=DP$, the pre-quantization activation-noise surrogate becomes
\begin{equation}
\boxed{
F_A^{\mathrm{grp}}(d)
=\frac{1}{12\kappa_A^2}
\sum_{G\in\mathcal G_A}
\left(
\sum_{t=1}^{m}
\max_{k\in G}\frac{X_{tk}^2}{d_k^2}
\right)
\left(
\sum_{i\in G}d_i^2\|P_{i,:}\|_2^2
\right).
}
\label{eq:grouped-activation-objective}
\end{equation}
The first factor measures the activation range seen by a group; the second
measures how much residual weight that group carries.  Their product explains
why reducing an activation outlier is useful only insofar as the corresponding
weight rescaling does not offset the gain.

For a singleton group $G=\{i\}$, $d_i^{-2}$ in the first factor cancels
$d_i^2$ in the second, so that group's contribution is independent of $d_i$.
For groups containing multiple channels, the maximum can switch between
channels while the residual energy changes continuously; smoothing then has
a genuine within-group effect.  The limiting case $G=\{1,\ldots,c\}$ gives
one activation scale per token.

The weight-noise channel has a parallel grouped model.  Let
$\mathcal G_W$ be the weight-quantization groups along the input dimension,
and let $\kappa_W=2^{b_W-1}-1$.  For output column $j$, define
\[
\delta_{W,j,G}(d)
=\frac{1}{\kappa_W}
\max_{k\in G}d_k|P_{kj}|,
\qquad G\in\mathcal G_W.
\]
With $h_i=(H_\perp)_{ii}$, the uniform-noise surrogate after eliminating
$L_2$ is
\begin{equation}
\boxed{
F_W^{\mathrm{grp}}(d)
=\frac{1}{12}
\sum_{j=1}^{n}\sum_{G\in\mathcal G_W}
\delta_{W,j,G}^2(d)
\sum_{i\in G}\frac{h_i}{d_i^2}.
}
\label{eq:grouped-weight-objective}
\end{equation}
Only the diagonal of $H_\perp$ appears here because the assumed weight errors
are uncorrelated within a group.  The off-diagonal entries remain relevant to
the actual GPTQ solve below.

\subsection{Geometric-Program Formulation}
\label{subsec:smoothing-gp}

For fixed $A$ and $B$, minimize the modeled two-channel objective
\[
\min_{d_i>0}
F_W^{\mathrm{grp}}(d)
+\lambda_A F_A^{\mathrm{grp}}(d),
\qquad \lambda_A\ge 0.
\]
Introduce positive epigraph variables $t_{j,G}$ for weight steps and
$s_{t,G}$ for activation steps.  An equivalent geometric program is
\begin{align}
\min_{d,t,s>0}\quad
&\frac{1}{12}
\sum_{j=1}^{n}\sum_{G\in\mathcal G_W}
t_{j,G}^{2}
\sum_{i\in G}h_i d_i^{-2}
\nonumber\\
&\quad
+\frac{\lambda_A}{12}
\sum_{G\in\mathcal G_A}
\left(\sum_{t=1}^{m}s_{t,G}^{2}\right)
\left(\sum_{i\in G}d_i^2\|P_{i,:}\|_2^2\right)
\label{eq:smoothing-gp}\\
\text{subject to}\quad
&\frac{d_i|P_{ij}|}{\kappa_W t_{j,G}}\le 1
&& (i\in G),
\nonumber\\
&\frac{|X_{ti}|}{\kappa_A d_i s_{t,G}}\le 1
&& (i\in G),
\nonumber\\
&\prod_{i=1}^{c}d_i=1.
\nonumber
\end{align}
The first constraint applies to every output $j$ and weight group
$G\in\mathcal G_W$; the second applies to every token $t$ and activation
group $G\in\mathcal G_A$.
At an optimum, each epigraph variable can be reduced to its corresponding
maximum, recovering
\eqref{eq:grouped-activation-objective} and
\eqref{eq:grouped-weight-objective}.  All objective terms are posynomials
with nonnegative coefficients, the inequalities are monomial bounds, and the
normalization is a monomial equality.  Hence the change of variables
\[
u_i=\log d_i,
\qquad
\tau_{j,G}=\log t_{j,G},
\qquad
\xi_{t,G}=\log s_{t,G}
\]
turns the constraints into affine inequalities and the logarithm of the
objective into a convex log-sum-exp function.  The idealized fixed-branch
$D$-subproblem therefore has a global optimum, subject to the stated noise
model and any imposed bounds on $d$.

The normalization removes an otherwise arbitrary common scaling: replacing
every $d_i$ by $a d_i$ leaves both modeled error terms unchanged.  Equivalently,
one may impose $\sum_i u_i=0$.  If a practical implementation clips the
smoothing range, bounds $-\log d_{\max}\le u_i\le\log d_{\max}$ can be added
without losing convexity.  A finite number of gradient steps in log space is
an approximate solver for this geometric program; the global-optimum claim
applies to a converged solution of the modeled subproblem, not automatically
to a finite-step implementation or to the measured quantized-layer error.

\subsection{Closed-Form Relaxation}
\label{subsec:smoothing-closed-form}

A simpler initialization follows by replacing grouped maxima with an
ungrouped $\ell_p$ range surrogate.  Define
\[
\omega_i^p
=\sum_{j=1}^{n}|P_{ij}|^p,
\qquad
h_i=(H_\perp)_{ii},
\qquad p\ge 1,
\]
and consider
\begin{equation}
\widehat F_W(d)
=C
\left(\sum_{i=1}^{c}\omega_i^p d_i^p\right)^{2/p}
\left(\sum_{i=1}^{c}h_i d_i^{-2}\right),
\label{eq:smoothing-relaxation}
\end{equation}
where $C>0$ does not affect the minimizer.  Assuming $h_i>0$ and
$\omega_i>0$, differentiation of $\log\widehat F_W$ with respect to
$u_i=\log d_i$ gives
\[
\frac{2\omega_i^p d_i^p}
{\sum_k\omega_k^p d_k^p}
-
\frac{2h_i d_i^{-2}}
{\sum_k h_k d_k^{-2}}
=0.
\]
The ratio of the two denominators is common to every channel.  Therefore,
up to an overall scale,
\begin{equation}
\boxed{
d_i^{\star}
\propto
\left(\frac{h_i}{\omega_i^p}\right)^{1/(p+2)}.
}
\label{eq:smoothing-closed-form}
\end{equation}
Because the logarithm of \eqref{eq:smoothing-relaxation} is convex in $u$,
this stationary point is a global minimizer of the relaxation.  Normalize its
geometric mean to one before refinement.

For $p=2$,
\[
d_i^{\star}
\propto
\frac{h_i^{1/4}}{\|P_{i,:}\|_2^{1/2}}.
\]
If the low-rank branch is absent, then $h_i=H_{ii}=\|X_{:,i}\|_2^2$ and
$P=W$, yielding
\[
d_i^{\star}
\propto
\left(
\frac{\|X_{:,i}\|_2}{\|W_{i,:}\|_2}
\right)^{1/2}.
\]
The square-root rule is thus exact for this relaxation, while the grouped
maxima and activation-noise term motivate the subsequent refinement.

\subsection{The activation-aware metric and target}
For fixed $D,L_1$, the modeled activation penalty depends on $\widehat R$
but not $L_2$. Eliminating $L_2$ therefore gives
\[
J(\widehat R)=\frac{1}{2}\|\widetilde W-\widehat R\|_{H_\perp}^2
+\frac{1}{2}\lambda_A\|\widehat R\|_{\overline\Sigma_A}^2.
\]
Define $M=H_\perp+\lambda_A\overline\Sigma_A$.
Both summands are positive semidefinite, implying
$\operatorname{null}(M)\subseteq\operatorname{null}(H_\perp)$.
Consequently $H_\perp\widetilde W\in\operatorname{range}(M)$ and
$T=M^\dagger H_\perp\widetilde W$ satisfies $MT=H_\perp\widetilde W$.
Expanding the square yields
\[
J(\widehat R)=\frac{1}{2}\|\widehat R-T\|_M^2
+\frac{1}{2}\|\widetilde W\|_{H_\perp}^2-\frac{1}{2}\|T\|_M^2.
\]
Thus both the target and metric change when activation noise is included.
At $\lambda_A=0$, the metric is generally singular: an ordinary inverse
cannot replace the pseudoinverse without regularization.

\paragraph{What regularization changes.}
With a common loaded metric $M_\epsilon=M+\epsilon I$ and target
$T_\epsilon=M_\epsilon^{-1}H_\perp\widetilde W$, the corresponding
objective is $J(\widehat R)+\frac{1}{2}\epsilon\|\widehat R\|_F^2$, up to a
code-independent constant. The weight-error path instead quantizes an
initialized residual $R_0=\widetilde W-L_1L_{2,0}$ under a stabilized
$H_\perp$. Its added penalty is centered at $R_0$. In addition, group scales
are constructed from the chosen target. Consequently, equal continuous
undamped losses need not produce equal grids, greedy updates, or codes.

\paragraph{Exact activation error versus its surrogate.}
The exact error is $\widetilde XC-E_A\widehat R$ and includes the cross
term in Eq.~\eqref{eq:exact_squared_error}. Replacing deterministic
activation rounding with a zero-mean conditional noise model is the step
that removes this cross term in expectation. The deployed activation error
multiplies $\widehat R$, not the unquantized provisional residual. Our
post-code objective preserves that distinction. The diagonal range-based
covariance does not estimate off-diagonal correlations and is not an exact
model of four-bit rounding.

\section{Algorithm}
\begin{algorithm}[htbp]
\caption{Detailed layerwise calibration of \method{}}
\label{alg:hsvdquant_detailed}
\small
\begin{algorithmic}[1]
\Require Weight $W$, calibration activations $X$, protected rank $r$,
         symmetric INT4 weight quantizer $Q_W$,
         per-token group-wise INT4 activation quantizer $Q_A$,
         code mode, activation weight $\lambda_A$
\Ensure $D,L_1,L_2,\widehat R$

\State $H_0 \gets X^\top X/m$
\State Initialize best candidate state
\State $A_{\rm prev} \gets \varnothing$, $L_{2,\rm prev} \gets \varnothing$

\For{$k=1,2$}
    \If{$k=1$}
        \State $H_{\perp,\rm prev} \gets H_0$
        \State $P_{\rm prev} \gets W$
    \Else
        \State
        $H_{\perp,\rm prev}
        \gets
        H_0
        -
        H_0 A_{\rm prev}
        (A_{\rm prev}^{\top}H_0A_{\rm prev})^{\dagger}
        A_{\rm prev}^{\top}H_0$
        \State
        $P_{\rm prev}
        \gets
        W-A_{\rm prev}L_{2,\rm prev}$
    \EndIf

    \State Update $D$ from $(H_{\perp,\rm prev},P_{\rm prev},X)$
    using Equation~\eqref{eq:d_closed_form} and optional cached refinement

    \State
    $\widetilde X \gets XD^{-1}$,
    $\widetilde W \gets DW$,
    $\widetilde H \gets D^{-1}H_0D^{-1}$

    \State Estimate $\overline{\Sigma}_A$
    using Equation~\eqref{eq:sigma_model}

    \State Initialize $(L_1,L_{2,0})$ by the weighted rank-$r$ fit
    in Equation~\eqref{eq:weighted_init_solution}

    \State Form the deflated metric
    $\widetilde H_\perp$
    using Equation~\eqref{eq:deflation}

    \If{activation-aware code optimization is enabled}
        \State Form $(M_Q,T_Q)$ using Equation~\eqref{eq:mq_tq}
        \State Construct $\widehat R$
        by GPTQ on target $T_Q$ with metric $M_Q$
        using $Q_W$
    \Else
        \State
        $R_0 \gets \widetilde W-L_1L_{2,0}$
        \State Construct $\widehat R$
        by GPTQ on target $R_0$ with metric $\widetilde H_\perp$
        using $Q_W$
    \EndIf

    \State Refit $L_2$ using Equation~\eqref{eq:l2_opt}

    \State Evaluate the W4A4 output MSE on cached calibration inputs
    in FP32 and retain the best visited state

    \State
    $A_{\rm prev} \gets D^{-1}L_1$,
    $L_{2,\rm prev} \gets L_2$
\EndFor

\State \Return best $(D,L_1,L_2,\widehat R)$

\end{algorithmic}
\end{algorithm}

\FloatBarrier
\section{Evaluation on Accelerated Diffusion Backbones}
\label{app:semantics}
\rev{Table~\ref{tab:semantic_complete} reports quantitative W4A4 results on accelerated few-step diffusion backbones: the 4-step FLUX.1-schnell and 4-step Qwen-Image Lightning variants, complementing the standard base model evaluations in Table~\ref{tab:dit_main}.
Teacher fidelity and standalone quality measure different properties: the method with the best PSNR need not have the highest CLIP or ImageReward score. Thus these evaluations capture distinct trade-offs between teacher fidelity and generation metrics across few-step regimes.}
\begin{table}[!htbp]
\centering
\fontsize{8}{9.5}\selectfont
\setlength{\tabcolsep}{2pt}
\renewcommand{\arraystretch}{1.05}
\caption{\textbf{W4A4 fidelity and generation quality on accelerated diffusion backbones (FLUX.1-schnell and Qwen-Image Lightning).} Both models use 4-step sampling on the predescribed dataset.
CLIP and ImageReward (IR) are absolute scorer means, not teacher-relative differences;
neither measures teacher fidelity.
Bold and underline mark the best and second-best unrounded means per subset and metric.
\method{} use group-64 residuals and the stated protected rank; DiRotQ uses matched A4/group-64.}
\label{tab:semantic_complete}
\begin{tabular*}{\linewidth}{@{\extracolsep{\fill}}lrrrrrrrr@{}}
\toprule
& \multicolumn{4}{c}{MJHQ} & \multicolumn{4}{c}{sDCI} \\
\cmidrule(lr){2-5}\cmidrule(lr){6-9}
Method & PSNR$\uparrow$ & LPIPS$\downarrow$ & CLIP$\uparrow$ & IR$\uparrow$ & PSNR$\uparrow$ & LPIPS$\downarrow$ & CLIP$\uparrow$ & IR$\uparrow$ \\
\midrule
\multicolumn{9}{@{}l}{\textbf{FLUX.1-schnell (4 steps)}} \\[1pt]
\textbf{H-SVDQuant $r=4$} & \underline{17.93} & \underline{0.260} & \underline{25.238} & 0.9029 & \underline{16.79} & \underline{0.278} & \underline{25.838} & 1.0761 \\
SVDQuant $r=4$ & 16.18 & 0.350 & \textbf{25.394} & 0.8951 & 15.12 & 0.363 & 25.791 & 1.0844 \\
SVDQuant $r=32$ & 17.31 & 0.298 & 25.223 & \textbf{0.9394} & 16.23 & 0.311 & 25.712 & \underline{1.0947} \\
DiRotQ & \textbf{18.56} & \textbf{0.241} & 24.994 & \underline{0.9089} & \textbf{17.41} & \textbf{0.257} & 25.588 & \textbf{1.0951} \\
OrbitQuant & 17.21 & 0.297 & 25.144 & 0.9007 & 16.23 & 0.306 & 25.836 & 1.0413 \\
ViDiT-Q & 14.74 & 0.467 & 24.758 & 0.7872 & 13.85 & 0.489 & \textbf{26.142} & 1.0235 \\
\midrule
\multicolumn{9}{@{}l}{\textbf{Qwen-Image Lightning (4 steps)}} \\[1pt]
\textbf{H-SVDQuant $r=4$} & \textbf{15.33} & \underline{0.328} & 29.422 & \textbf{1.6268} & \underline{13.33} & \underline{0.391} & 25.541 & \textbf{1.2425} \\
SVDQuant $r=4$ & 14.31 & 0.382 & \textbf{29.639} & 1.5958 & 12.84 & 0.437 & 25.508 & 1.2173 \\
SVDQuant $r=32$ & 15.09 & 0.339 & 29.580 & 1.6063 & 13.30 & 0.406 & \textbf{25.602} & 1.2372 \\
DiRotQ & \underline{15.18} & \textbf{0.327} & \underline{29.597} & \underline{1.6160} & \textbf{13.50} & \textbf{0.383} & \underline{25.543} & \underline{1.2400} \\
\bottomrule
\end{tabular*}
\end{table}

\FloatBarrier

\section{Qualitative Results of Qwen-Image}
\label{app:qwen_qualitative}
\begin{figure}[H]
\centering
\includegraphics[width=0.96\linewidth]{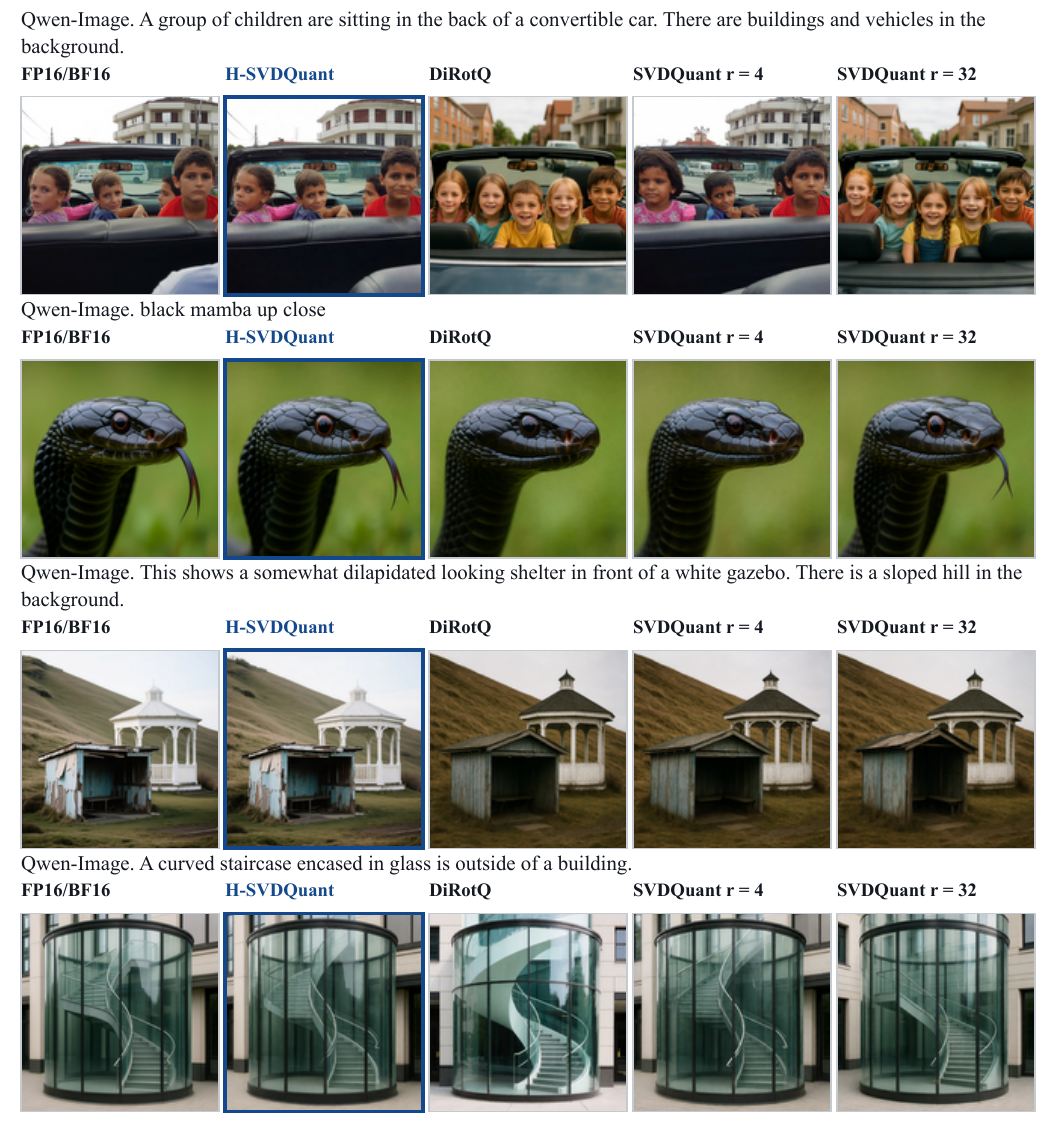}
\caption{\textbf{Qwen-Image base-model reconstructions.}
The first column is the high-precision reference; the blue outline identifies
\method{}. All methods share the displayed prompt within each row. These
qualitative examples complement the complete-subset results in
Table~\ref{tab:dit_main}.}
\label{fig:qual_qwen_base}
\end{figure}

\begin{figure}[H]
\centering
\includegraphics[width=0.92\linewidth]{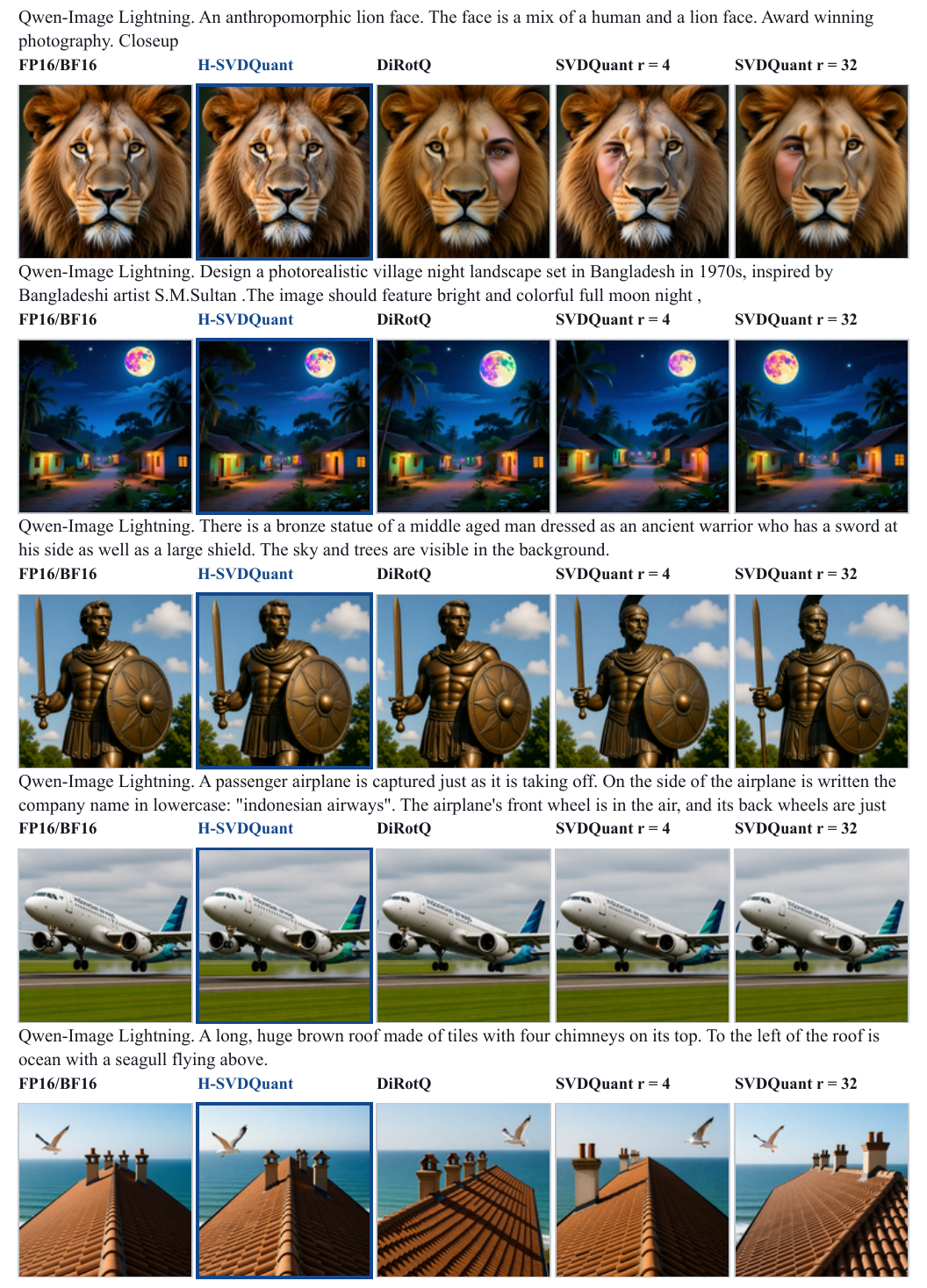}
\caption{\textbf{Four-step Qwen-Image Lightning reconstructions.}
Column order follows Figure~\ref{fig:qual_qwen_base}.
The accelerated quantitative comparison appears in
Table~\ref{tab:semantic_complete}.}
\label{fig:qual_qwen_lightning}
\end{figure}

Figures~\ref{fig:qual_qwen_base} and~\ref{fig:qual_qwen_lightning}
show representative qualitative reconstructions for the 50-step base model and four-step Lightning variant, respectively, evaluated under identical seeds and prompts against DiRotQ and both SVDQuant ranks.

\FloatBarrier
\clearpage

\section{Full Paired Prompt Distributions}
\label{app:paired_distributions}
\begin{figure}[htbp]
\centering
\includegraphics[width=\linewidth]{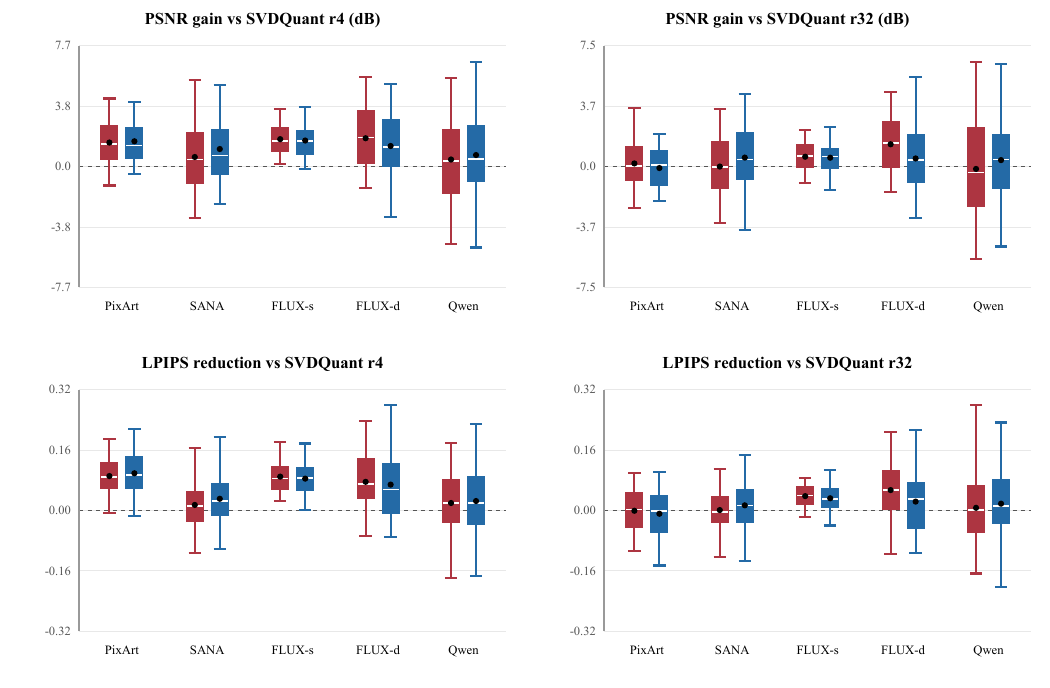}
\caption{\textbf{Paired teacher-fidelity differences across five diffusion backbones.}
This expanded version of Figure~\ref{fig:intro_method_pair} adds Qwen-Image
and compares rank-4
\method{} against SVDQuant ranks 4 and 32 on identical prompts and seeds.
Columns separate baseline rank; rows show PSNR gain and LPIPS reduction.
Boxes show the interquartile range, white bars the medians, whiskers the
5th--95th percentiles, and black dots the means. Red denotes MJHQ and blue
sDCI; positive values favor \method{}. The compared checkpoints retain
their respective activation quantizers and are complete-recipe comparisons.
Qwen-Image images are paired by index under identical prompt, seed, and teacher protocols.}
\label{fig:paired_medium}
\end{figure}
\FloatBarrier



\section{Between-block and within-block propagation pilot}
\label{app:block}
\begin{table}[htbp]
\centering\small
\caption{\textbf{Three-way calibration replay pilot.} All rows use rank 4,
W4A4/group-64, $\lambda_A=0$, and the weight-error code path. Within each
model, evaluation uses the same 16 MJHQ prompts and image identifiers.
These pilot checkpoints are not substituted into Table~\ref{tab:dit_main}.}
\label{tab:propagation}
\begin{tabular}{lllrr}
\toprule
Model & Between blocks & Within block & PSNR & LPIPS\\
\midrule
PixArt-$\Sigma$ & quantized & sequential & 15.523 & 0.4834\\
& reference & sequential & 15.701 & 0.4675\\
& reference & independent & 15.833 & 0.4662\\
\midrule
FLUX-schnell & quantized & sequential & 17.471 & 0.3003\\
& reference & sequential & 17.703 & 0.2961\\
& reference & independent & 17.059 & 0.3151\\
\bottomrule
\end{tabular}
\end{table}
Relative to reference/sequential, the two PixArt PSNR differences are
$-0.177$~dB ($95\%$ paired interval $[-0.839,0.412]$) and
$+0.133$~dB ($[-0.239,0.529]$). This pilot does not establish a ranking.
For FLUX-schnell, the corresponding differences are $-0.232$~dB
($[-0.500,0.016]$) and $-0.644$~dB ($[-1.395,-0.003]$);
both LPIPS intervals include zero. Despite worse image fidelity in the
independent FLUX run, its largest local module MSE is slightly smaller.
Local errors are measured on each route's own calibration activations and
are not paired errors on a shared held-out activation set. The pilot therefore
illustrates a limitation of local MSE as a predictor of full-image behavior.

\end{document}